\documentclass{ut-thesis}
\usepackage[colorlinks]{hyperref} % for links
\usepackage[backend=bibtex]{biblatex} % for references
\usepackage{graphicx} % for embedding graphics
\usepackage{booktabs} % for pretty tables
\usepackage{amsfonts} % for pretty tables
\usepackage{amsmath}
\usepackage{lipsum} % for gibberish text
\usepackage{algorithm}
\usepackage{algpseudocode}
\author{Jimmy Lin}
\title{On the Interaction Between Differential Privacy and Gradient Compression in Deep Learning}
\degree{Master of Science}
\department[]{School of Computer Science}
\gradyear{2022}
\begin{document}
  \frontmatter
    \maketitle
    \begin{abstract}
      While differential privacy and gradient compression are separately well-researched topics in machine learning, the study of interaction between these two topics is still relatively new. We perform a detailed empirical study on how the Gaussian mechanism for differential privacy and gradient compression jointly impact test accuracy in deep learning. 

The existing literature in gradient compression mostly evaluates compression in the absence of differential privacy guarantees, and demonstrate that sufficiently high compression rates reduce accuracy. Similarly, existing literature in differential privacy evaluates privacy mechanisms in the absence of compression, and demonstrates that sufficiently strong privacy guarantees reduce accuracy.  

In this work, we observe while gradient compression generally has a negative impact on test accuracy in non-private training, it can sometimes improve test accuracy in differentially private training. Specifically, we observe that when employing aggressive sparsification or rank reduction to the gradients, test accuracy is less affected by the Gaussian noise added for differential privacy. These observations are explained through an analysis how differential privacy and compression effects the bias and variance in estimating the average gradient.

We follow this study with a recommendation on how to improve test accuracy under the context of differentially private deep learning and gradient compression. We evaluate this proposal and find that it can reduce the negative impact of noise added by differential privacy mechanisms on test accuracy by up to 24.6\%, and reduce the negative impact of gradient sparsification on test accuracy by up to 15.1\%.
    \end{abstract}
    % \begin{dedication}
    % \end{dedication}
    % \begin{acknowledgements}
    % \end{acknowledgements}
    \tableofcontents
    \listoftables
    \listoffigures
  \mainmatter
    \chapter{Introduction}
\section{Motivation}
As applications of machine learning techniques grew in domains such as robotics, healthcare, and advertisement, so too has the need for protecting the privacy of users contributing training data. Federated learning \cite{McMahan2017} served as a first step towards privacy by keeping participating data distributed across privately managed devices and only sharing gradient vectors during training. However, Deep Leakage from Gradient \cite{Zhu2020} demonstrates the insufficiency of this approach by reconstructing private data samples using gradient vectors shared amongst participants during training. DPSGD \cite{Abadi2016}, an application of the differentially private Gaussian mechanism \cite{Dwork2014} to deep learning, offers an optimization framework that implements and quantifies differential privacy bounds in the context of federated learning. Since Deep Leakage from Gradient relies on information leaked through gradients and DPSGD protects privacy by adding noise to gradients, DPSGD serves as a good way to protect against reconstruction attacks.

Federated learning involves frequent communication of gradient vectors from data owners to a central parameter server. These communications occur across wide area networks which generally have very limited upstream network capacity for the data owners. As deep learning models grow in size and complexity, the size of gradient vectors can become too large to be suitable for sharing over wide area networks. Gradient compression algorithms can alleviate this problem when they correctly assume some properties of gradient vectors such as sparsity \cite{Lin2017,Xiao2021} or low rank \cite{Vogels2019}. Most algorithms operate on the assumption that gradient vectors can be represented in a subspace of much lower dimensionality than the parameter space used to represent the model. However, the addition of noise in DPSGD violates this assumption.

In this work, our goal is to characterize how the combination of DPSGD and gradient compression interact and produce a joint impact on the test accuracy for deep learning models. We do so by measuring the effect of gradient noise on accuracy at varying compression rates, and similarly, the effect of compression on accuracy at varying levels of noise. In particular, we observe the following:

\begin{itemize}
    \item The addition of noise tends to decrease accuracy, and this effect is more pronounced for larger models than smaller ones.
    \item Increasing the compression rate in non-private training tends to decrease accuracy, and this effect is more pronounced in image-classification models than text classification models.
    \item In the presence of noise, gradient compression in smaller models can sometimes recover some of the accuracy lost to noise.
    \item The use of aggressive gradient compression when training smaller models can result in a reduced sensitivity to gradient noise.
\end{itemize}

We explain these observations by analyzing the error which DPSGD and compression introduces to the average gradient estimation. In particular, we observe the following:
\begin{itemize}
    \item Reduced accuracy in training can be explained through mean-squared error estimating the average gradient.
    \item The error is mostly made up of variance, and a small amount of bias.
    \item The variance component of the error is mostly introduced by the noise addition in the Gaussian mechanism as part of implementing differential privacy guarantees.
    \item Both sparsification and rank reduction leads to a large reduction of variance in exchange for a small amount of bias, but leading to an overall decrease in mean-squared error, hence it can lessen the reduction in test accuracy in private training.
\end{itemize}

\section{Background}
In this section, we describe the definition of differential privacy mechanisms and gradient compression algorithms used in our study.

\subsection{Differential Privacy}
To test the effects of differential privacy, we adopt the Opacus \cite{Yousefpour2021} library. It implements a combination of the Gaussian Mechanism \cite{Dwork2014} for ($\epsilon$, $\delta$)-differentially private queries, DPSGD \cite{Abadi2016} for differentially private deep learning, and Renyi differential privacy accountant \cite{Mironov2017} for privacy accounting.

($\epsilon$, $\delta$)-differential privacy \cite{Dwork2014} is defined so: Let function $\mathcal{M}: \mathcal{X} \rightarrow \mathcal{Y}$ be a mapping from domain $\mathcal{X}$ to range $\mathcal{Y}$. Define the adjacency of to any two sets of data $d, d' \in \mathcal{X}$ to mean that $max(|d - d'|, |d' - d|) \le 1$ (i.e. $d$ and $d'$ differ by at most a single sample.) $\mathcal{M}$ is defined to be ($\epsilon$, $\delta$)-differential privacy if it follows the following statement

\begin{equation}
\begin{split}
    \forall\: d, d' \in \mathcal{X} \textrm{ s.t. } max(|d - d'|, |d' - d|) \le 1 \\
    \forall\: \mathcal{S} \subseteq \mathcal{Y} \\
    \textrm{Pr}[\mathcal{M}(d) \in \mathcal{S}] \leq e^{\epsilon}\textrm{Pr}[\mathcal{M}(d') \in \mathcal{S}] + \delta
\end{split}
\end{equation}

An intuitive understanding of this definition is: any observation made about the output $\mathcal{M}(d) \in \mathcal{S}$ can be alternatively explained by $\mathcal{M}(d') \in \mathcal{S}$ with a lower-bounded probability. Giving $d$ (and by symmetry, $d'$) plausible deniability when an output in $\mathcal{S}$ is observed. The likelihood of any alternative explanation $\textrm{Pr}[\mathcal{M}(d') \in \mathcal{S}]$ is lower-bounded relative to original explanation $\textrm{Pr}[\mathcal{M}(d) \in \mathcal{S}]$ the bounded by a factor $e^{\epsilon}$ and constant $\delta$. As $\epsilon$ and $\delta$ approaches 0, the likelihood of alternative explanations become just as likely the original explanation and perfect anonymity is guaranteed.

To guarantee that any arbitrary mapping $\mathcal{M}$ satisfies such an inequality, the most widely accepted implementation is to replace $\mathcal{M}$ with a new function $\mathcal{M'}$ that applies the following post-processing steps to the outputs of $\mathcal{M}$

\begin{equation}
    \mathcal{M'}(x) = \textrm{project}_{C}(\mathcal{M}(x)) + \mathcal{N}(0, \sigma C)
\end{equation}

The first step is to project all outputs of $\mathcal{M}$ into a spherical region centered at the origin with a bounded radius $C \ge 0$. All projections are then perturbed by adding independently sampled noise from a Gaussian distribution $\mathcal{N}(0, \sigma C)$ with a standard deviation $\sigma C$ proportional to $C$. The proportionality factor $\sigma$ controls the level of privacy guarantee available per observation, with larger values guaranteeing higher privacy. The resulting $\mathcal{M'}$ satisfies an infinite set of ($\epsilon$, $\delta$)-differential privacy where $\delta \geq \frac{4}{5}e^{\frac{-(\sigma\epsilon)^2}{2}}$ and $\epsilon < 1$. This particular post-processing is referred to as a Gaussian Mechanism in the differential privacy literature. The choice of $C$ doesn't impact the privacy guarantee, however it does effect the accuracy of training. In our work we tried a few different values for each task and picked the one that gives the best accuracy after a single epoch. More advanced strategies exist such as adaptive clipping \cite{Andrew2022} work which proposes the an adaptive clipping radius $C$ dynamically set to a differentially private estimate of a fixed quantile of the gradient norms. 

Traditionally, the Gaussian mechanism is applied to queries on a data base. DPSGD \cite{Abadi2016} introduced this method to deep learning by applying it to each gradient computation with respect to each individual data sample in the training data set. This guarantees a quantifiable amount of privacy between participating data samples and, by extension, the individuals supplying those data samples. To account for the accumulating privacy cost during training, they implement a privacy accountant to keep track of the increasing values of $\epsilon$.

The definition of differential privacy can be changed in many ways. For example, defining the set of data belonging to an individual user as a units to anonymize \cite{McMahan2018}, as opposed to individual samples being treated as units.

\subsection{Gradient Compression}
In this work we focus on two different approaches to gradient compression: Deep Gradient Compression \cite{Lin2017} and PowerSGD \cite{Vogels2019}

Deep Gradient Compression is an algorithm that produces a layer-wise sparse representation of the gradient vector by representing only the elements with relatively larger magnitudes in each layer. The unrepresented elements are not communicated, with the receiver interpreting them as zeros. This algorithm further compresses all represented elements by removing the low-order bits in their floating point representation. The receiver also interprets the removed bits to be zero.

This algorithm works under the assumption that most elements of a gradient vector are close to zero, and that the loss function is smooth. Under this assumption, approximating many near-zero values as zero produces a permissibly small change to the model update which produces a permissibly small change in loss for a smooth loss function. There exists other variants of sparsification such as one using an entropy-based criteria \cite{Xiao2021} for selecting coordinates to approximate as zero. However, we focus on using Deep Gradient Compression as an example of sparsification.

PowerSGD is an algorithm that reshapes the layer-wise gradient vector into a square matrices and learns a low-rank factorization of these matrices. The resulting factors are communicated in lieu of the original matrices when it would result in a lower bandwidth usage. This algorithm works under the assumption that the rows of each square matrix are coordinates that span a much smaller set of dimensions than the number of columns in the square matrix. Under this assumption, an approximate low-rank factorization of the square matrix can be produced and used to reconstruct the square matrix without large amounts of error. 

Notably, for this assumption to be true, there would have to be high correlation between the coordinates of the original gradient vector, which implies an approximate low-rank representation for all gradient vectors. This is similar to the assumption of near-zero values in gradient vectors, however the assumption is generalized to assume near-zero projections along a large set of directions in the parameter space. This reveals that rank reduction can be viewed as a non-axis-aligned generalization of sparsification. Due to this connection between sparsification and rank reduction, both sparsification and rank reduction tend to bias the gradients towards the origin and reduce their variance..

Coordinate-wise quantization is another approach to gradient compression however, it is usually not used in isolation since its compression rate is upper-bounded by 64 (floating point values are generally represented in 64 bits, and the minimum coordinate-wise representation is 1 bit.) For this reason, coordinate-wise quantization is not the most competitive approach in literature. While it is possible quantization may interact with differential privacy differently than sparsification and rank reduction, we leave this direction as an area for future study.

\section{Related Work}
In this section, we discuss related works which involve both differential privacy and gradient compression.

There exists a growing body of work that is focused on improving the efficiency of differential privacy and compression. These include testing the effectiveness of various compression algorithms in differentially private training. DP-SCAFFOLD \cite{Noble2021} applies the work of SCAFFOLD \cite{Karimireddy2019} to DPSGD, and find that the control variates designed to reduce the impact of non-IID data partitions can also reduce the variance introduced by differential privacy mechanisms. Q-DPSGD \cite{Ding2021} explores the effectiveness of gradient quantization applied before and after the Gaussian mechanism and benchmarks it to be computationally faster than SDM-DSGD \cite{Zhang2021} which applies a randomized unbiased sparsification after the Gaussian mechanism. FL-CS-DP \cite{Kerkouche2020} explores the use of compressive sensing, where they view the gradient vector as a time series that can be transformed into frequency space, keeping only the low-frequency values. They propose a novel formulation of the compression optimization to improve upon traditional DCT (Discrete Cosine Transform) compression.

The works above attempt to find combinations of differential privacy and compression mechanisms that achieve the greatest resource efficiency, with the resource being time, bandwidth, or privacy budget. Our work's main focus is to offer insight on the relationship between compression, differential privacy, and accuracy. We hope that these insights inspire novel ideas that result in greater resource efficiency.

There also exists a number of works that explore compression mechanisms which already introduce noise. These mechanisms can be modified to provide differential privacy guarantees on top of the pre-existing compression capabilities. Count-Sketch \cite{Li2019,Chen2021} is one such mechanism that inherently introduces randomness through random hash functions. Dithered quantization \cite{Amiri2021} is another approach which adds noise before quantization. MVU \cite{Chaudhuri2022} adds this noise after quantization by sampling from discrete distribution. They also formulation an optimization that minimizes the distribution variance while satisfying differential privacy guarantees.

These work explore the potential of re-purposing pre-existing randomness in compression algorithms towards differential privacy. Similarly, their goal is to prove the effectiveness of this approach against a baseline, less so to provide a deep analysis of how their compression interacts with differential privacy mechanisms.

Additionally, there is research in differential privacy and compression in contexts other than deep learning such as data base queries \cite{Zhou2009}. While the same privacy and compression mechanisms can often used across many contexts, their interaction can be dependent on the type of information being protected and compressed. The assumptions one can make regarding gradient vectors in deep learning can't be generally made about arbitrary data base queries. We study the specific context of deep learning in hopes of finding unique insights that would otherwise be hidden in a more general context.

    \chapter{Methods}
In this section, we describe the tasks, models, and hyperparameter settings used to conduct our experiments. We also define some metrics used in our results. Refer to the following repository for an exmaple of how to run these experiments: \url{https://github.com/Jimmy-Lin/privacy-ml-systems}

\section{Tasks and Models}
To evaluate the generality of our insight across different tasks in deep learning, we train 4 models on 4 different tasks: Surnames, CIFAR-10, SNLI, CIFAR-100

\subsection{Surnames Task}
The goal of this task is to classify the language associated with an alphabetic surname, given a choice of 18 different languages. We train with a learning rate of 2.0 and a batch size of 32 for 100 epochs. Refer to the following URL to find a copy of the data set \url{https://github.com/spro/practical-pytorch/tree/master/data/names.}

The model we train is a 256-character-set LSTM model with a single LSTM layer of 64 embedding dimensions and 128 output dimensions, followed by a fully connected layer of 18 classes and a softmax activation. Refer to \cite{Hochreiter1997} for details on the LSTM cell architecture.

\subsection{CIFAR-10 Task}
The goal of this task is to classify the object at the centre of a 32x32 coloured image, given a choice of 10 different object classes. We train with a learning rate of 0.1 and a batch size of 128 for 100 epochs. Refer to \cite{Krizhevsky2009} for more information on this data set.

The model we train is a 3-block CNN, each block containing a biased convolution layer of kernel size 3 and stride 1. Each convolution is followed by an instance normalization with a momentum value of 0.1, a ReLU activation, an average pooling with a pool size of 2, and a spatial dropout with probability 0.1. The 3 blocks differ only by their number of output filters: 32, 64, and 128. After the 3 blocks, we follow with 2 biased hidden layers of 256 and 512 units respectively, each using ReLU activation and dropout with probability 0.25. Lastly, the model finishes with a fully connected layer into 10 classes and a softmax activation.

\subsection{SNLI Task}
The goal of this task is to classify the logical relation between a pair of English sentences, the relation can be either "entailment", "contradiction", or "neutral". We train with a learning rate of 0.05 and a batch size of 32 for 1 epoch. Refer to ~\cite{Bowman2015} for more information on this data set.

We fine-tune a pre-trained "bert-based-case" model which can be found at \url{https://huggingface.co/bert-base-cased}. We freeze all parameters except for the classifier, pooling, and final layer of the encoder. Refer to ~\cite{Devlin2018} for details on the BERT architecture.

\subsection{CIFAR-100}
The goal of this task is to classify the object at the centre of a 32x32 coloured image, given a choice of 100 different object classes. We train with a learning rate of 1.0 and a batch size of 64 for 100 epochs. Refer to \cite{Krizhevsky2009} for more information on this data set.

We train a modified version of ResNet-18. Specifically, we set the global average pooling that follows the stack of residual blocks to output 2x2 channels instead of 1x1. We find this modification results in much better accuracy on this data set. Refer to ~\cite{He2015} for details on the ResNet architecture.

\section{Differential Privacy and Compression Hyperparameter Settings}
In this section, we describe how we configure the differential privacy mechanism and compression algorithms.

\subsection{Differential Privacy Mechanism Settings}
For the Surnames task, we use a clipping radius of 3.0 and a $\delta$ value of 0.00008. For the CIFAR-10 task, we use a clipping radius of 5.0 and a $\delta$ value of 0.00001. For the SNLI task, we use a clipping radius of 21.0 and a $\delta$ value of $\frac{1}{549361}$. For the CIFAR-100 task, we use a clipping radius of 1000.0 and a $\delta$ value of 0.00001. To vary the privacy level, we use noise multiplier values of 0.0, 0.4, and 0.8. Note that we clip the gradients even in the non-private training so that the changes in privacy guarantee is solely attributed to the noise addition. Gradient clipping in non-private training is common practice for the purpose of limiting the impact of exploding gradients.

Clipping radius is selected based on an approximate median of the gradient norm at the first iteration. This is based on the observation of adaptive clipping \cite{Andrew2022} that clipping approximately 50\% of the gradients in a batch appear to work well. However, we keep a fixed value instead of adjusting it over the course of training. While this selection may be unlikely in practice, it serves as a good way to standardize across tasks that exhibit different gradient norms. We select $\delta$ values by picking a value roughly on the order of $\frac{1}{n}$ where $n$ is the number of training samples in the training data set. This is the recommended upper bound on $\delta$ \cite{Dwork2014} for ($\epsilon$, $\delta$)-differential privacy in literature.

\subsection{Compression Algorithm Settings}
To vary the compression rate of Deep Gradient Compression (DGC) we configure the DGC algorithm to compression rates of $1$, ${16}$, and ${256}$. To vary the compression rate of PowerSGD, we configure the PowerSGD algorithm to use approximation ranks of $1$ and $16$. In practice, PowerSGD doesn't offer low compression rates as approximation ranks above 16 tend to incur severely large compute overhead. Due to this computational overhead and the large size of layers in ResNet, we set the approximation ranks to $1$ only for the CIFAR-100 tasks instead of $1$ and $16$.

\section{Metric Definitions}

\subsection{Accuracy}
We measure the test accuracy of a model at the end of every epoch and use a average of the last 10 epochs to represent the final model accuracy. In the case of tasks which train for only 1 epoch, we simply take the single measurement of test accuracy.

\subsection{Bandwidth Usage}
Since upstream network capacity is generally far more scarce in wide area networks than downstream network capacity, we measure only the upstream network usage which consists of mainly the gradient vectors uploaded to parameter servers per client. We assume all vectors are transmitted in COO format when estimating the number of bytes sent.

\subsection{Privacy Bound}
For measurements of privacy bound, we defer to the Opacus library's implementation of Renyi differential privacy accountant to track the $\epsilon$ value. We fix $\delta$ as a hyperparameter and quantify differences in privacy guarantee solely through $\epsilon$. Smaller values of $\epsilon$ indicate a stronger differential privacy guarantee.

    \chapter{Results}
In this section, we discuss the results of training each task at different combinations of differential privacy guarantees, compression algorithms, and compression rates. We acknowledge that higher test accuracy may be achievable through state-of-the-art architectural designs. Since the goal of this study is to characterize the relationship between different configurations and well-studied architectures, it is unnecessary to find the most optimal architecture and configuration for each data set.

\section{Effects on Test Accuracy}
In this section, we focus on the effects we observe on test accuracy.

\subsection{Surnames Task}
\begin{figure}[h]
    \centering
    \includegraphics[width=0.8\linewidth]{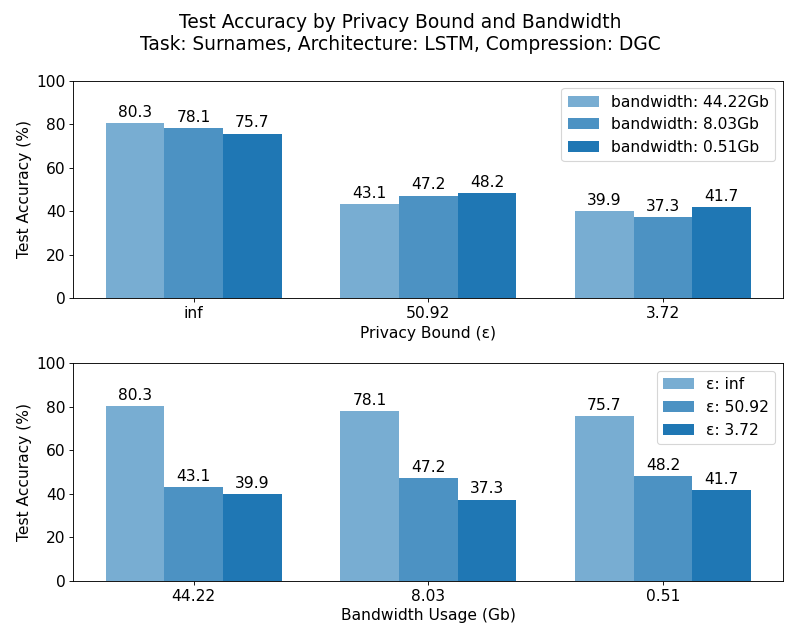}
    \caption{Test accuracy averaged over last 10 epochs after 100 epochs of training the Surnames task using DGC. (Top) grouped by differential privacy bound ($\epsilon$). (Bottom) grouped by upstream network usage (Gb).}
    \label{fig:skyline_lstm_local_dgc}
\end{figure}
\begin{figure}[h]
    \centering
    \includegraphics[width=0.8\linewidth]{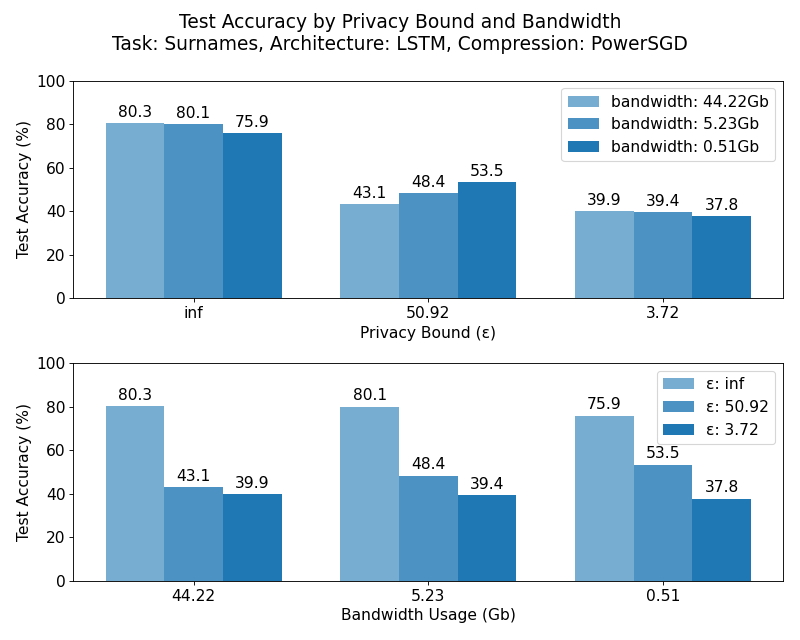}
    \caption{Test accuracy averaged over last 10 epochs after 100 epochs of training the Surnames task using PowerSGD. (Top) grouped by differential privacy bound ($\epsilon$). (Bottom) grouped by upstream network usage (Gb).}
    \label{fig:skyline_lstm_local_power_sgd}
\end{figure}
Figure \ref{fig:skyline_lstm_local_dgc} (top) shows the accuracy measurements grouped by their differential privacy bound ($\epsilon$). While most groups experience a small decrease in accuracy of a few percent, we observe that the group with $\epsilon = 50.92$ experiences a slight increase in accuracy of 5.1\%.
Figure \ref{fig:skyline_lstm_local_dgc} (bottom) shows the accuracy measurements grouped by their upstream network usage (Gb). Within each group we see a large decrease in accuracy when the noise multiplier is increased. The uncompressed group which used 44.22Gb experiences an accuracy drop of 40.4\%. The compressed group experiences an accuracy drop of 34.0\%.

In figure \ref{fig:skyline_lstm_local_power_sgd}, we observe very similar patterns to figure \ref{fig:skyline_lstm_local_dgc}. Remarkably, the accuracy increase in the group with $\epsilon = 50.92$ is even larger at a 10.4\% increase.

Overall we observe that this task is relatively robust to gradient compression, losing only a few percent in accuracy. Surprisingly, an increase is accuracy is sometimes observed with increasing compression rate. This observation was more noticeable when using the PowerSGD compression

\subsection{CIFAR-10 Task}
\begin{figure}[h]
    \centering
    \includegraphics[width=0.8\linewidth]{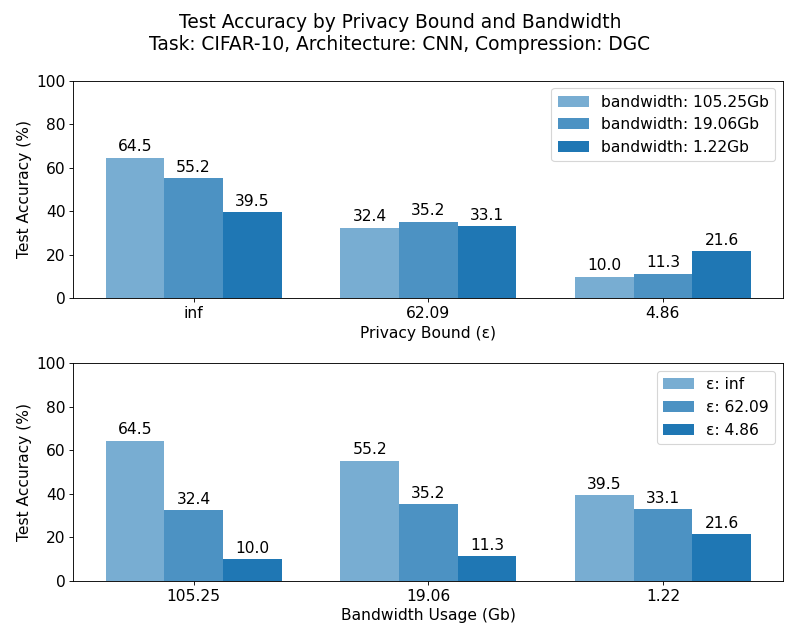}
    \caption{Test accuracy averaged over last 10 epochs after 100 epochs of training the CIFAR-10 task using DGC. (Top) grouped by differential privacy bound ($\epsilon$). (Bottom) grouped by upstream network usage (Gb).}
    \label{fig:skyline_cnn_local_dgc}
\end{figure}
\begin{figure}[h]
    \centering
    \includegraphics[width=1.0\linewidth]{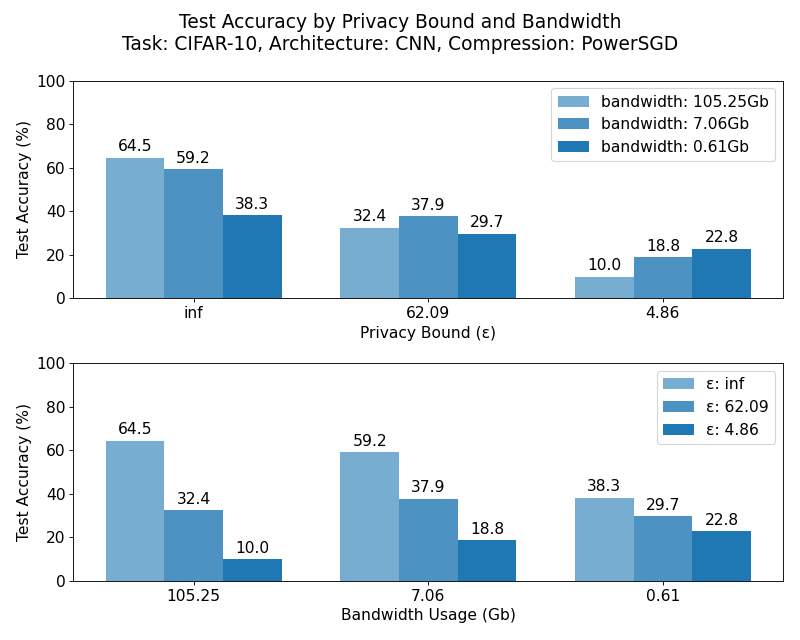}
    \caption{Test accuracy averaged over last 10 epochs after 100 epochs of training the CIFAR-10 task using PowerSGD. (Top) grouped by differential privacy bound ($\epsilon$). (Bottom) grouped by upstream network usage (Gb).}
    \label{fig:skyline_cnn_local_power_sgd}
\end{figure}

Figure \ref{fig:skyline_cnn_local_dgc} (top) shows the accuracy measurements grouped by their differential privacy bound ($\epsilon$). We observe that this task is noticeably more sensitive to compression than the Surnames task. Compression can decrease the accuracy by as much as 25.0\%. Once again, we observe an increase in accuracy with higher compression rate, but this time it occurs within the group with $\epsilon = 4.86$ and the increase in accuracy is 11.6\%.
Figure \ref{fig:skyline_cnn_local_dgc} (bottom) shows the accuracy measurements grouped by their upstream network usage (Gb). We notice that the in-group range of accuracy decreases as the amount of bandwidth used decreases. This starts at a range of 54.5\% (bandwidth = 105.25Gb) to a range of 17.9\% (bandwidth = 1.22Gb).

In figure \ref{fig:skyline_cnn_local_power_sgd}, we observe very similar patterns to figure \ref{fig:skyline_cnn_local_dgc}.

Overall we observe that this task is relatively more sensitive to compression than the Surnames task. The the increase in accuracy in private training when compression rate is increased is observed again, similar to what we observe in the Surnames task. This time the increase is similar between compression algorithms

\subsection{SNLI Task}
Figure \ref{fig:skyline_bert_local_dgc} (top) shows the accuracy measurements grouped by their differential privacy bound ($\epsilon$). We observe that this task, in the non-private $\epsilon = \infty$ case, is very robust to gradient compression. It is inconclusive whether this can be said about the private training cases, since the models produced are of similar accuracy to a random prediction. This is due to this task being very sensitive to the noise added by the differential privacy mechanism. We do see a very slight increase in accuracy of 1.9\% when $\epsilon is 0.89$, however this is not a very significant amount.
Figure \ref{fig:skyline_bert_local_dgc} (bottom) shows the accuracy measurements grouped by their upstream network usage (Gb). In every group, the accuracy lost due to noise is the dominant factor in changes to accuracy. We do see a slight decrease in sensitivity to noise from 48.9\% to 44.7\% but this amount is not very conclusive.

In figure \ref{fig:skyline_bert_local_power_sgd}, we observe very similar patterns to figure \ref{fig:skyline_bert_local_dgc}.

We observe that this task is very sensitive to noise, with it being the dominant factor in observable loss in accuracy. We do observe the increase in accuracy correlated with compression and a reduction in noise sensitivity when compression is added. However, the amount is much smaller this time and it is hard to use this as conclusive evidence. We attribute this to the fact that noise is so dominant in it's effect on accuracy for this task.

\subsection{CIFAR-100 Task}
Figure \ref{fig:skyline_resnet_local_dgc} (top) shows the accuracy measurements grouped by their differential privacy bound ($\epsilon$). We observe that this task is very sensitive to noise and compression. Similar to the SNLI task, the model's accuracy is no better than random prediction when noise is added by the differential privacy mechanism. In the non-private case, we see a 22.2\% decrease in accuracy after compression. Figure \ref{fig:skyline_resnet_local_dgc} (bottom) shows the accuracy measurements grouped by their upstream network usage (Gb). We observe that noise dominates the decrease in accuracy in the non-compressed group with bandwidth = 417.5Gb. However, compression also plays a role in decreasing the accuracy in non-private training.

In figure \ref{fig:skyline_resnet_local_power_sgd}, we observe very similar patterns to figure \ref{fig:skyline_resnet_local_dgc}.

We observe that this task is very sensitive to noise, but also compression. No interesting pattern can be observed from the experiments ran on this task, due to most trials resulting in minimal accuracy.

\subsection{General Observations}
We observe that the larger models used in the SNLI and CIFAR-100 tasks are more sensitive to noise than the smaller models. While smaller models such as the LSTM and CNN do experience loss of accuracy due to noise, they aren't immediately rendered par with random predictions. Additionally, the image classification tasks are noticeably more sensitive to compression than the text classification tasks. This could be attributed to the data type being classified, or potentially common architectural components in image classification vs text classification (eg. convolution, normalization, pooling vs embedding, LSTM, self-attention).

We observe that in the tasks involving smaller models (Surnames and CIFAR-10), at some levels of differential privacy guarantees (finite $\epsilon$ value), increasing the compression rate can increase the model accuracy. We also observe that compressing the gradient appears to reduce the sensitivity of accuracy to noise. We hypothesize that compression has a way of reducing the negative impact of noise.

In the tasks involving larger models (SNLI, CIFAR-100), we observe either a very weak form of the trend or no such trend at all. We attribute this to their relatively higher sensitivity to noise.

\section{Convergence Analysis}
In this section, we analyze the changes in test accuracy over the course of training which is measured after ever epoch. We omit the SNLI task from this analysis since it is trained for only 1 epoch, and thus has no further information to show when visualized as a time series.

\subsection{Surnames Task}
Figure \ref{fig:timeline_lstm_local_dgc} shows the progression of test accuracy over the course of training the Surnames task. We observe that all trials actually reach their plateau within the first 10 epochs. The accuracy of private training trials exhibit very large variability over time, but their compressed counterparts appear to reduce this variability. Finally, the non-compressed, non-private training trial loses test accuracy after an initial peak within the first 10 epochs. It's compressed counterpart doesn't exhibit this behaviour but it also doesn't exceed it in final accuracy.

Figure \ref{fig:timeline_lstm_local_dgc_smooth} provides a smoothed view of the time series for better comparison of the private training trials. To achieve this smoothing, we use a mean convolution over the time axis with width 20 and no padding at end points.

The same observations can be made in \ref{fig:timeline_lstm_local_power_sgd_smooth} and \ref{fig:timeline_lstm_local_power_sgd} when using the PowerSGD compression algorithm.

\subsection{CIFAR-10}
\begin{figure}[h]
    \centering
    \includegraphics[width=0.8\linewidth]{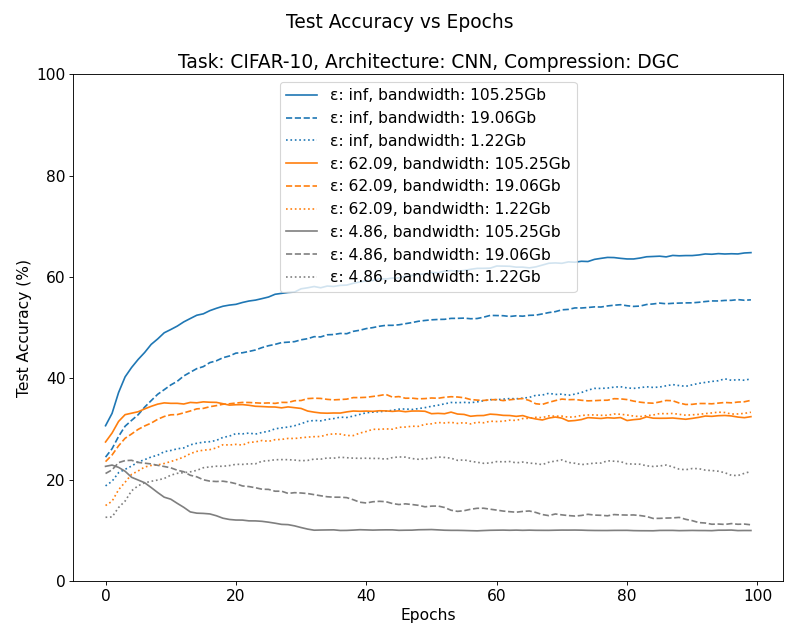}
    \caption{Test Accuracy over 100 Epochs of the CIFAR-10 task.}
    \label{fig:timeline_cnn_local_dgc}
\end{figure}

Figure \ref{fig:timeline_cnn_local_dgc} shows the progression of test accuracy over the course of training the CIFAR-10 task. We observe that the variability in test accuracy over time is consistently small for all trials. This time non-compressed trials in private training are the ones that decrease in test accuracy after an initial peak within the first 10 epochs. Furthermore, we find that this drop in accuracy appears to explain why the non-compressed trials show lower accuracy than their compressed accuracy in figure \ref{fig:skyline_cnn_local_dgc}. When we look at the accuracy in the first 10 epochs, the accuracy is higher when compression is lower. When we look at the last 10 epochs, the accuracy is higher when the compression is higher.

We see a similar effect when applying the PowerSGD compression algorithm in figures \ref{fig:timeline_cnn_local_power_sgd} and \ref{fig:skyline_cnn_local_power_sgd}.

\subsection{CIFAR-100}
Figures \ref{fig:timeline_resnet_local_dgc} and \ref{fig:timeline_resnet_local_power_sgd} show the progression of test accuracy over the course of training the CIFAR-100 task. We observe  that the variability over time is very small for this task as well, and accuracy is mostly increasing steadily over the course of training (if increasing at all).

\subsection{General Observations}
We observe that the Surnames task show much higher variability in accuracy over time when noise is added. Additionally, trials not using compression sometimes experience a peak early in training, followed by gradual loss of accuracy. This effect is lessened by compression. In some instances, this leads to the non-compressed trial finishing with lower accuracy than the compressed trial.

\section{Gradient Error}
In this section, we analyze the correlation between gradient error and test accuracy and break down the error to better understand what contributes to our observed decrease in accuracy. We define gradient error as the mean squared error between a gradient vector average prior to an application of the differential privacy mechanism and gradient compression and it's counterpart after the differential privacy mechanism and gradient compression. Specifically, we measure the gradient error at the beginning of training.

We target empirical average gradient as follows: Let $\{\textbf{X}_i, \textbf{y}_i\}_{i=1}^B$ denote a set of $B$ input-output pairs $(\textbf{X}_i, \textbf{y}_i)$ randomly sampled from the training data samples. Let $L$ denote a loss function we wish to optimize with respect to $\theta$, an $m$-dimensional vector of parameters. We define the $m$-dimensional vector $\textbf{g}$ as the empirical average gradient over $B$ training data samples as a target we wish to estimate through possibly noisy and/or biased samples.
\begin{equation}
    \textbf{g} = \frac{1}{B}\sum_{i=1}^{B}\nabla_{\theta}L(\textbf{X}_i, \textbf{y}_i)
\end{equation}
We define a mechanism $F: \mathbb{R}^{b,m} \Rightarrow \mathbb{R}^m$ as a function that takes as input the set of $B$ gradient samples and outputs an estimate of $\hat{\textbf{g}}$. This mechanism is allowed to be stochastic. We view the composition of our differential privacy mechanism and gradient compression as one such mechanism.
\begin{equation}
    \hat{\textbf{g}} = F(\{\nabla_{\theta}L(\textbf{X}_i, \textbf{y}_i)\}_{i=1}^B)
\end{equation}
Since $F$ can be stochastic, it may produce a different estimate each time. For this reason, we measure the mean-squared difference between an estimate $\hat{\textbf{g}}$ and the target $\textbf{g}$ for $n$ independent instances of $\hat{\textbf{g}}$. We measure mean-squared error of the average gradient estimate as follows:
\begin{equation}
    MSE(F, \textbf{g}) = \frac{1}{n}\sum_{i=1}^n||\hat{\textbf{g}}_i - \textbf{g}||_2^2
\end{equation}

In the results that follow we use $n = 100$ as the sample size for estimating the gradient error.

\subsection{Correlation between Gradient Error and Test Accuracy}
In this section, we show scatter plots between test accuracy and gradient error for each task and compression algorithm. 

\begin{figure}[h]
    \centering
    \includegraphics[width=0.8\linewidth]{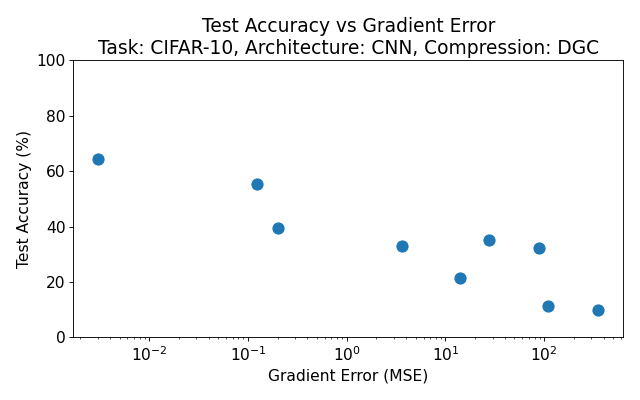}
    \caption{Test accuracy vs gradient error (log base-10 scale) for the CIFAR-10 task with DGC.}
    \label{fig:explanation_cnn_local_dgc}
\end{figure}

In figure \ref{fig:explanation_cnn_local_dgc}, we observe a trend that decrease in test accuracy coincides with increase in gradient error. The gradient error has been plotted on a log scale to better illustrate this. This correlation exists in all other tasks which we demonstrate in figures \ref{fig:explanation_lstm_local_dgc} to \ref{fig:explanation_resnet_local_power_sgd}.

\subsection{Effects on Gradient Error}
In this section, we analyze how gradient error relates to the level of noise added by differential privacy mechanisms and gradient compression algorithms.
\begin{figure}[h]
    \centering
    \includegraphics[width=0.8\linewidth]{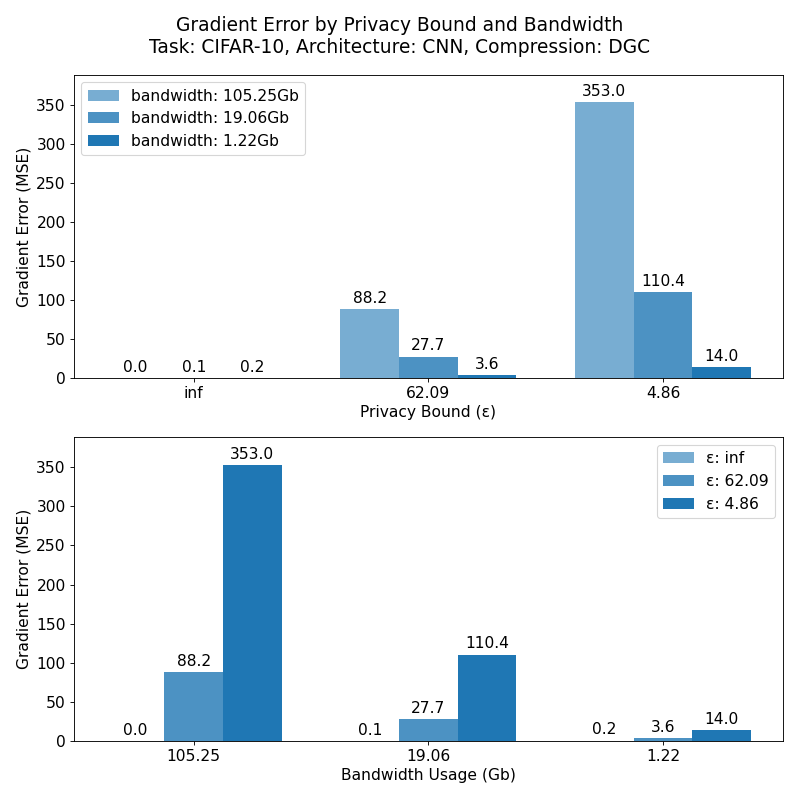}
    \caption{Gradient Error vs Privacy Bound ($\epsilon$) and Bandwidth (Gb) for the CIFAR-10 task with DGC.}
    \label{fig:error_skyline_cnn_local_dgc}
\end{figure}

In figure \ref{fig:error_skyline_cnn_local_dgc}, we observe that the gradient error is mostly contributed by the differential privacy mechanism's addition of noise (up to 353.0 units). While the compression algorithm can contribute to gradient error (up to 0.2 units), it often reduces the gradient error already contributed by the differential privacy mechanism (from 353.0 units down to 14.0 units). We believe this to be related to the correlation between compression rate and test accuracy in private training of small models. This can also be observed in other tasks in figures \ref{fig:error_skyline_lstm_local_dgc} to \ref{fig:error_skyline_resnet_local_power_sgd}.

Specifically, if compression reduces gradient error, and lower gradient error correlates with higher test accuracy, then it isn't unreasonable that compression correlates with higher test accuracy through the lowering of gradient error. 

We do see the same error reduction in large models, but no increase in test accuracy. It's possible that the error reduction is simply not strong enough to overcome the effect of noise.

\subsection{Gradient Error Breakdown: Bias vs Variance}
In this section, we analyze a breakdown of the gradient error into bias and variance. 

\begin{figure}[h]
    \centering
    \includegraphics[width=0.8\linewidth]{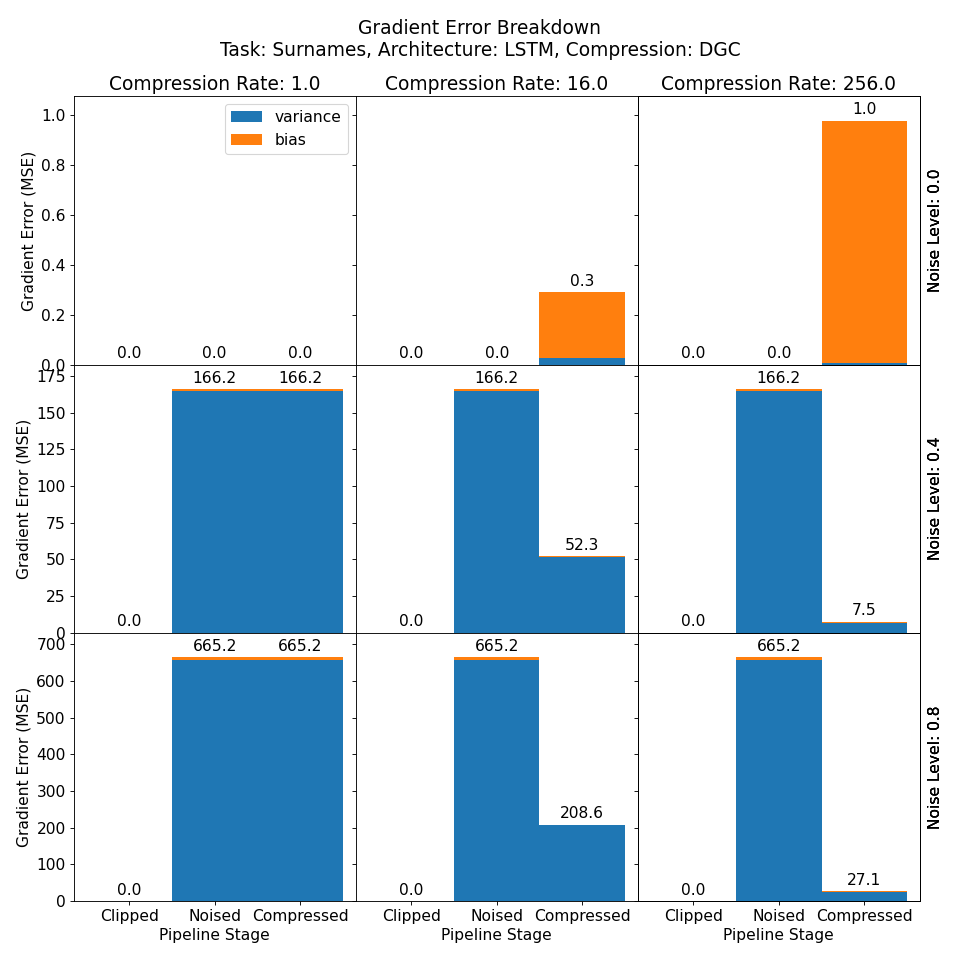}
    \caption{Gradient Error vs Privacy Bound ($\epsilon$) and Bandwidth (Gb) for the Surnames task with DGC.}
    \label{fig:error_breakdown_lstm_local_dgc}
\end{figure}

\begin{figure}[h]
    \centering
    \includegraphics[width=0.8\linewidth]{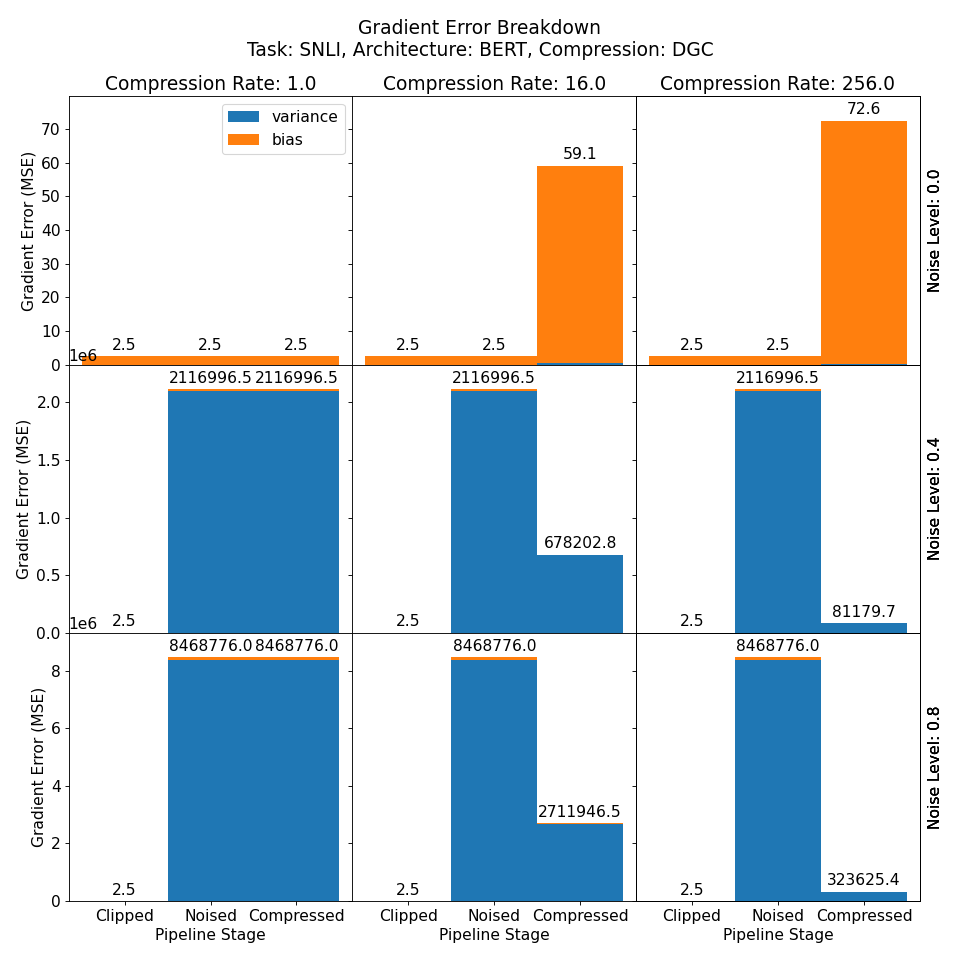}
    \caption{Breakdown of gradient error into bias and variance at different stages (after clipping, after noising, and after compression) for the SNLI task with DGC.}
    \label{fig:error_breakdown_bert_local_dgc}
\end{figure}

It is well known that the expected squared error of an estimator, in this case the expected gradient, can be decomposed into a bias component and a variance component. With sufficiently large $n$, this is also true of the empirical average of squared error.
\begin{equation}
    \mathbb{E}\bigl[\:||\:\hat{\textbf{g}} - \textbf{g}\:||_2^2\:\bigr] = ||\:\mathbb{E}\bigl[\hat{\textbf{g}}\bigr] - \textbf{g}\:||_2^2 + \mathbb{E}\bigl[\:||\:\hat{\textbf{g}} - \mathbb{E}\bigl[\hat{\textbf{g}}\bigr]\:||_2^2\:\bigr]
\end{equation}

The bias component is simply the deviation caused by the mechanism $F$, and the variance is the variability across different instances of estimates due to the stochasticity of $F$.

Here we observe through figures \ref{fig:error_breakdown_lstm_local_dgc}, \ref{fig:error_breakdown_bert_local_dgc} the following: Clipping (left bar in each subplot) at the current configuration introduces relatively minimal gradient error, and when it does it tends to introduce bias (orange) not variance (blue). Noising (middle bar in each subplot) introduces a very significant amount of error and when it does it is overwhelmingly variance (blue) and not bias (orange). Compression (right bar in each subplot) introduces some amount of bias (orange) but not variance (blue) and it is relatively small compared to the variance introduced by noising. Furthermore, compression reduces the variance introduced by noising. This is also supported in other tasks shown in figures \ref{fig:error_breakdown_lstm_local_power_sgd} to \ref{fig:error_breakdown_resnet_local_power_sgd}.

A high-level way of interpreting this is that the two compression algorithms we tested introduce a bias towards the origin, but in doing so they reduce the variance of our average gradient estimation. In the context of differentially private training where noise contributes a large amount of variance, reducing a large amount of variance in exchange for a small amount of bias can reduce the overall error when estimating the average gradient. It can be said that the compression has a regularizing effect on our estimation of the average gradient.

\section{Optimizing the Bias-Variance Trade-Off}
In this section, we show the suboptimality of selecting clipping values based on the 50th percentile of gradient norms. We demonstrate why it makes sense to drastically reduce the clipping value for larger models, and that the clipping value has an optimal value related to the shrinkage coefficient of the James-Stein estimator \cite{James1961}.

\subsection{Minimal-Error Clipping}
In this section, we empirically test the clipping value that minimizes gradient error as well as provide an approximate theoretical model for guessing the optimal clipping value.

\begin{figure}[h]
    \centering
    \includegraphics[width=0.8\linewidth]{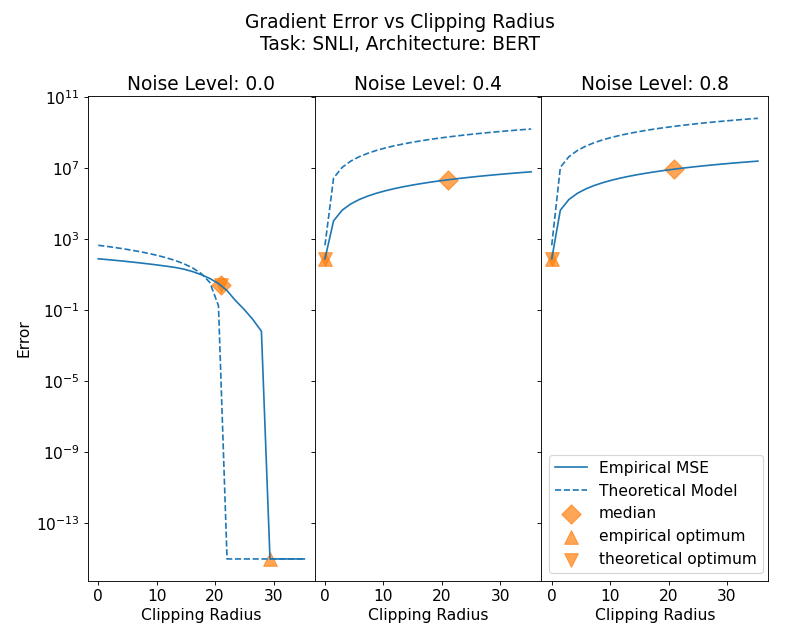}
    \caption{Relationship between gradient error and clipping value for SNLI task.}
    \label{fig:optimal_clipping_bert}
\end{figure}
\begin{figure}[h]
    \centering
    \includegraphics[width=0.8\linewidth]{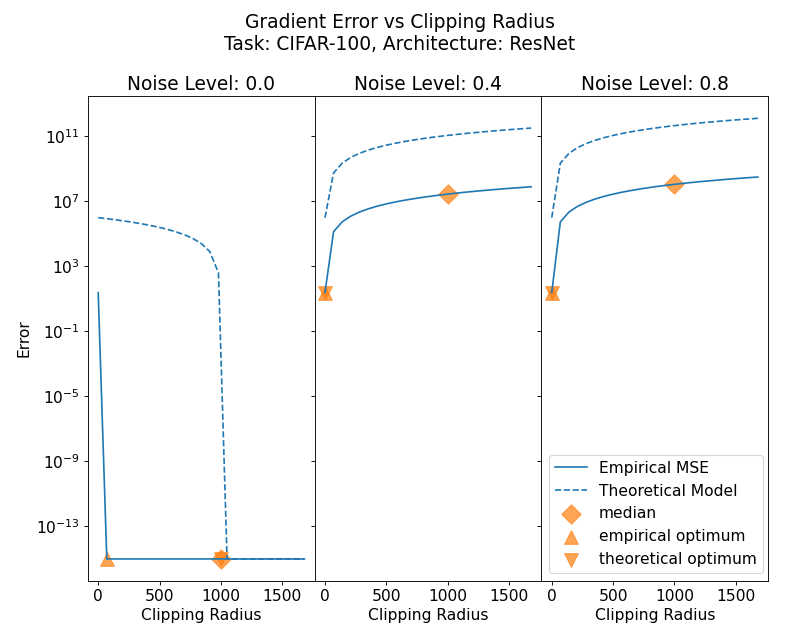}
    \caption{Relationship between gradient error and clipping value for CIFAR-100 task.}
    \label{fig:optimal_clipping_resnet}
\end{figure}

Figures \ref{fig:optimal_clipping_bert} and \ref{fig:optimal_clipping_resnet} demonstrate that the strategy of setting the clipping radius to be the median of gradient norms \cite{Andrew2022} does not minimize the gradient error. We propose the following theoretical model of the gradient error. More examples of this are shown in figures \ref{fig:optimal_clipping_lstm} and \ref{fig:optimal_clipping_cnn}.

\begin{equation}
    \textrm{Approximate Error} = max(0,(||\textbf{g}||_2 - C)^2) + mC^2\sigma^2
\end{equation}

We further simplify this model into a convex differential form as follows to allow us to directly solve for the minimum by differentiating with respect to $C$.

\begin{equation}
    \textrm{Differentiable Approximate Error} = (||\textbf{g}||_2 - C)^2 + mC^2\sigma^2
\end{equation}

\begin{equation}
    C^* = \textrm{argmin}_C \bigl( (||\textbf{g}||_2 - C)^2 + mC^2\sigma^2 \bigr) = \frac{||\textbf{g}||}{1 + m\sigma^2}
\end{equation}

Shown in figures \ref{fig:optimal_clipping_bert}, and \ref{fig:optimal_clipping_resnet}, the minimum for this model generally leads to at least a couple orders of magnitude of decrease in gradient error. Of course, the quality of such a model depends on the knowledge of the median gradient norm $||\textbf{g}||_2$. The advantage this model provides is a better utilization of the knowledge of $||\textbf{g}||_2$, should it be available either exactly or approximately with differential privacy guarantees in the case of \cite{Andrew2022}. It is a better strategy than simply setting $C = ||\textbf{g}||_2$, as it takes into account the effect of dimensionality $m$ and the noise multiplier $\sigma$. More examples of this are shown in figures \ref{fig:optimal_clipping_lstm} and \ref{fig:optimal_clipping_cnn}.

Worth noting in figure \ref{fig:optimal_clipping_resnet} (left), is that the assumed knowledge of the median gradient norm $||\textbf{g}||_2$ appears to be produce a bad theoretical model. This results in the theoretical minimum being much larger than the empirical model. We believe this is due to a large difference between the norm of the average gradient and the median of the gradient norms, and that perhaps the norm of the average gradient would be better at informing the theoretical model.

Another use for this model is providing efficient clipping values that are well-tuned across different levels of differential privacy. With the use of this model, it can be shown that our methodology of keeping the same clipping threshold for different levels of differential privacy is actually suboptimal. However, to our best knowledge, there's no work in the literature of differential privacy that currently suggests the optimal clipping threshold depends on the dimensionality and privacy level.

While it is possible to manually tune the clipping parameter with many repeated trials, this can be infeasible in the context of differentially private training. This is because training a model to convergence, even if it uses a suboptimal clipping value, can incur a large privacy cost. It is much more efficient to perform a differentially private query of the average gradient norm, and compute a reasonable clipping radius directly.

\begin{figure}[h]
    \centering
    \includegraphics[width=1.0\linewidth]{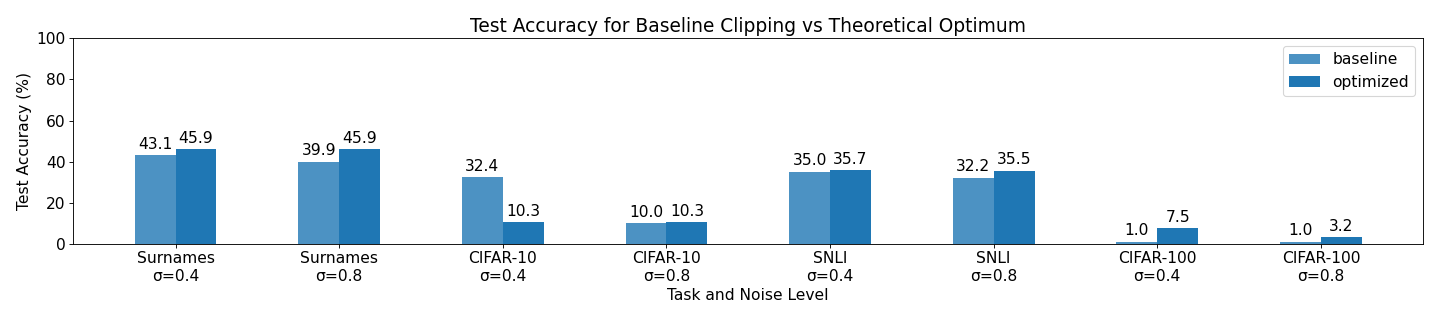}
    \caption{Result of optimizing clipping based on theoretical model}
    \label{fig:skyline_clipping}
\end{figure}

Figure \ref{fig:skyline_clipping} shows that the there is some benefit to optimizing the clipping value in most cases with gains of up to 6.5\% accuracy. However, not all tasks benefit from this. The CIFAR-10 task experienced worse accuracy after the change in clipping.

\subsection{Bias-Variance Optimization through Intermediate Processing}
In this section, we investigate the effectiveness reducing the error by processing the differentially private gradient estimates after the privacy mechanism as opposed to directly adjusting the privacy mechanism. Our proposed mechanism requires no prior knowledge about the gradients before the privacy mechanism. So there is no need to perform differentially private queries for information such as the median of gradient norms.

We propose this new algorithm which functions reduce the error introduced by noise addition and compression: Denoise. In this algorithm, both the sender and receiver keep track of a velocity term which is the an exponential average of the past average gradients. At each iteration, the sender updates it's velocity using the new average gradient. We define this change in velocity as acceleration. The sender then performs a top-k sparsification of both the acceleration and the velocity vector, and compares the norm of the error produced by sparsification in both cases. The sender sends either the sparse velocity or the sparse acceleration to the receiver along, whichever results in the least compression error, along with a single-bit flag to signal whether the message contains a velocity vector or acceleration vector. The sender accumulates a residual term to feed back into the next message, but the residual is decayed by some factor. The receiver updates it's velocity by replacing it with a new velocity vector or adding the acceleration vector. The key to our compression algorithm is the use of temporal averaging to reduce noise and the option to choose between either velocity or acceleration, one of which may incur less compression error at any given round. Details of this algorithm are written in \ref{alg:doubletake}

\begin{figure}[h]
    \centering
    \includegraphics[width=0.8\linewidth]{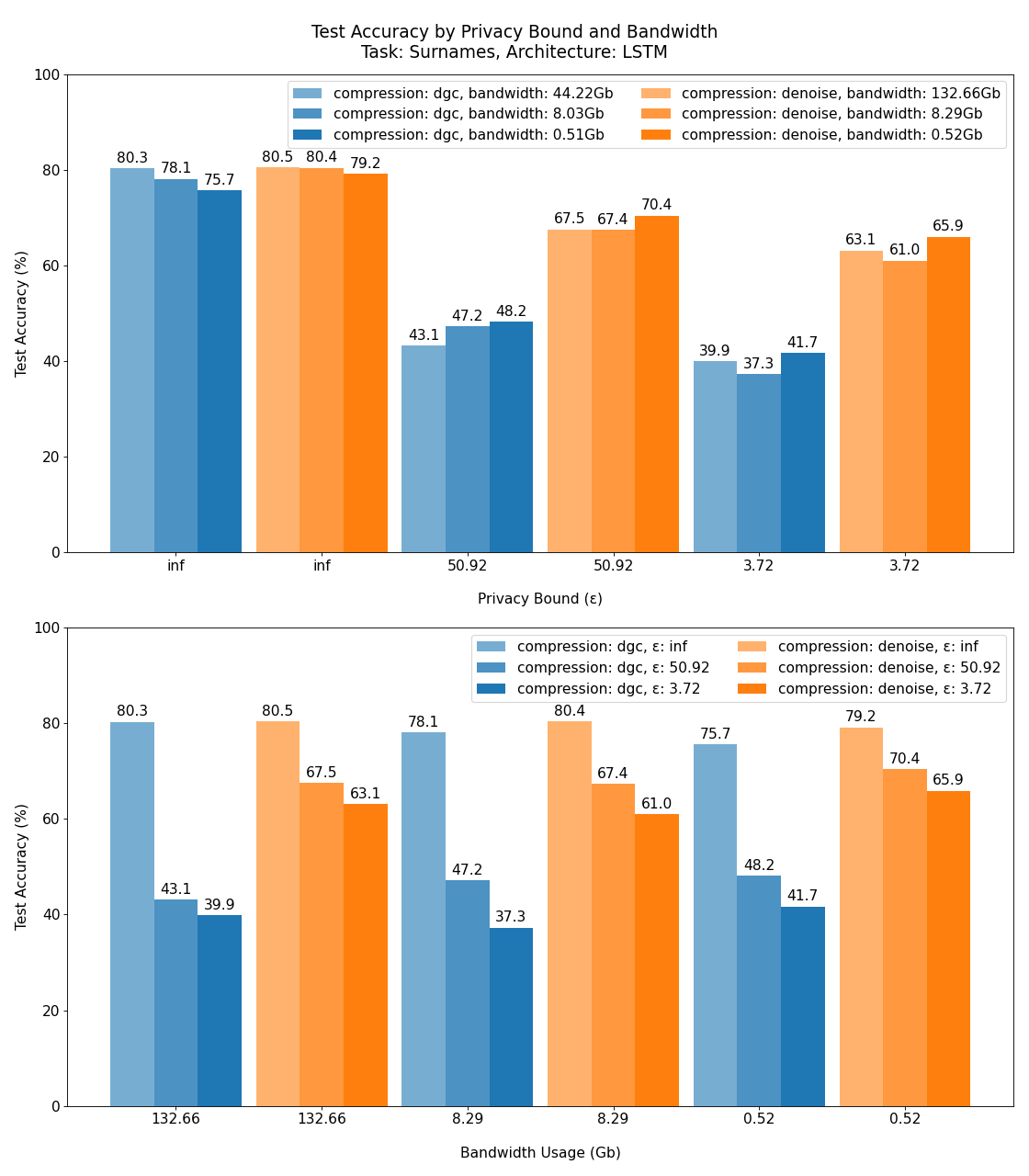}
    \caption{Effectiveness of introducing our algorithm to variance combinations of differential privacy guarantees and gradient compression (sparsification).}
    \label{fig:postprocess_lstm}
\end{figure}
\begin{figure}[h]
    \centering
    \includegraphics[width=0.8\linewidth]{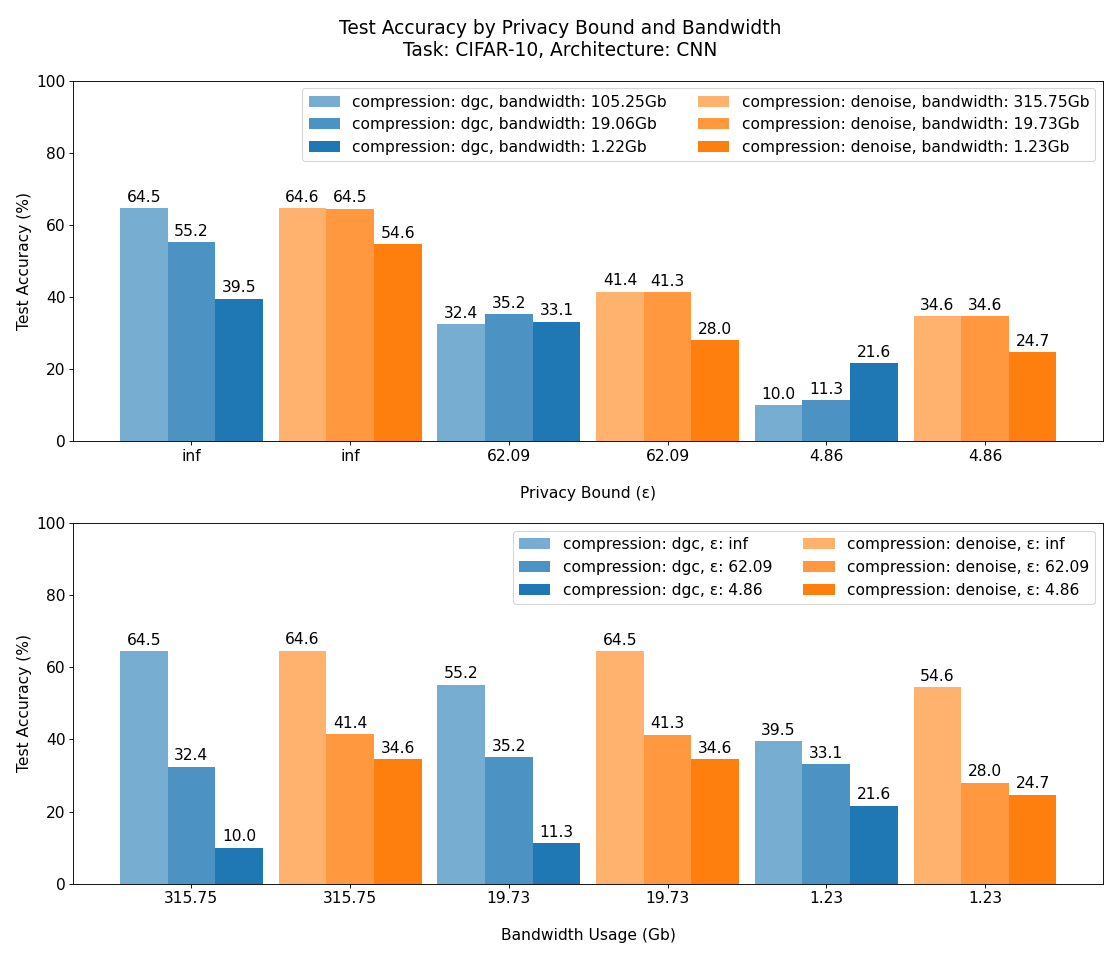}
    \caption{Effectiveness of introducing our algorithm to variance combinations of differential privacy guarantees and gradient compression (sparsification).}
    \label{fig:postprocess_cnn}
\end{figure}
\begin{figure}[h]
    \centering
    \includegraphics[width=0.8\linewidth]{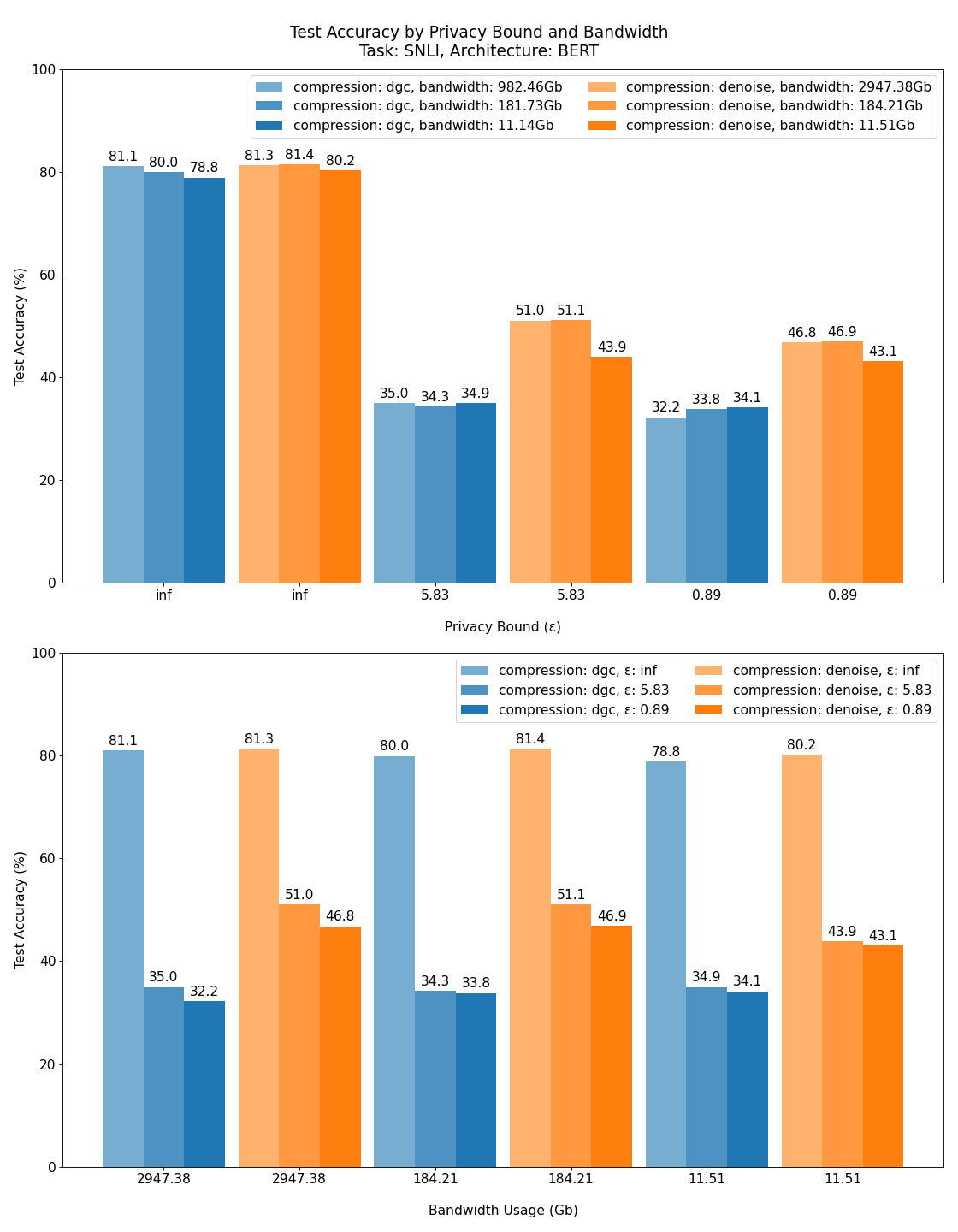}
    \caption{Effectiveness of introducing our algorithm to variance combinations of differential privacy guarantees and gradient compression (sparsification).}
    \label{fig:postprocess_bert}
\end{figure}
\begin{figure}[h]
    \centering
    \includegraphics[width=0.8\linewidth]{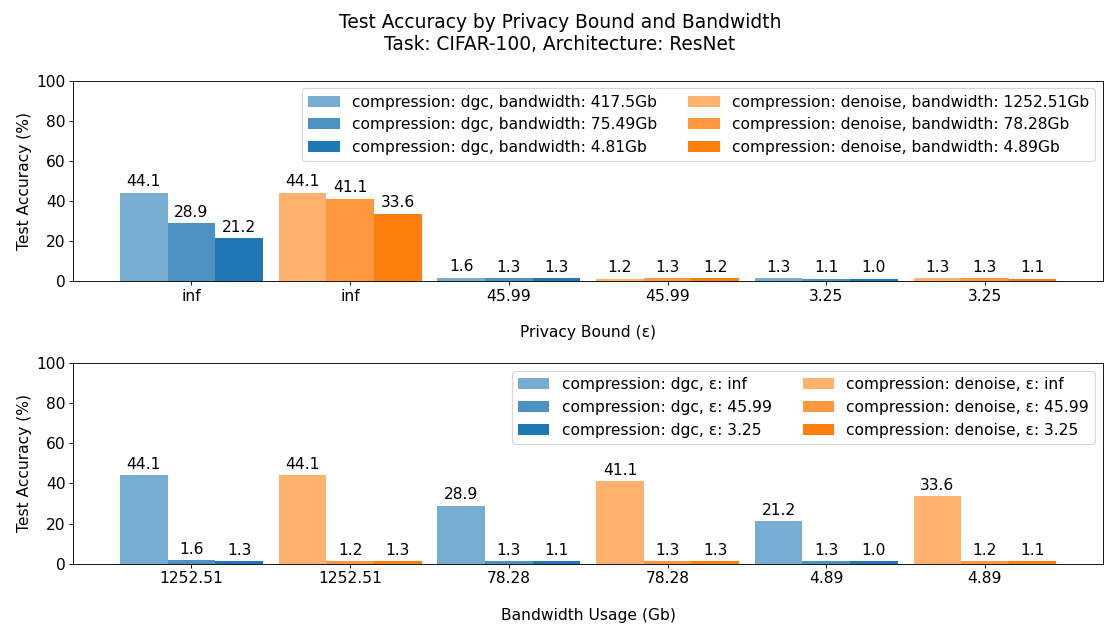}
    \caption{Effectiveness of introducing our algorithm to variance combinations of differential privacy guarantees and gradient compression (sparsification). This result was ran with a modified clipping radius of 70.0.}
    \label{fig:postprocess_resnet}
\end{figure}

Figures \ref{fig:postprocess_lstm} to \ref{fig:postprocess_resnet} show the impact of differentially private training with sparsification (DGC) compared to differentially private training with model-wise sparsification and denoise applied in between. 

In the Surnames, CIFAR-10, and SNLI tasks, the addition of denoise improves accuracy for every combination of privacy and compression except for CIFAR-10  with $\epsilon$ = 62.09 and bandwidth = 1.23Gb. 

In the non-private cases, we see that allowing the compression algorithm to choose between two different messages allows us to compress more aggressively with less reduction in accuracy (3.5\% less in Surnames, 15.1\% in CIFAR-10, 1.4\% in SNLI, and 12.4\% in CIFAR-100). In the non-compressed cases, we see that the second clipping reduces the variance enough to enable stronger differential privacy guarantees without as much reduction in accuracy (23.2\% in Surnames, 24.6\% in CIFAR-10, 14.6\% in SNLI, and 0.0\% in CIFAR-100). Unfortunately, this effect was not observable in the CIFAR-100, due to it being particularly sensitive to noise. 

In the CIFAR-100 task, we observe greater compressibility with denoise, but the noise sensitivity did not see observable differences. We show results of running with a lower clipping radius of 70.0 in figure \ref{fig:postprocess_resnet} (suggested as the empirical optimum from testing minimal-error clipping) but saw the same results.

The function of our algorithm is orthogonal to differential privacy and compression. For this reason, it can be used in conjunction with any gradient noising mechanism and any gradient compression algorithm, and is not in direct competition with the prior in either area.

\begin{algorithm}
\caption{Denoise: Intermediate processing which reduces variance and compression error}\label{alg:doubletake}
\textbf{Input:} $ \{\:\nabla_{\theta}L(\textbf{X}_i, \textbf{y}_i)\:\}_{i=1}^B $ \textit{Set of input gradients} \\
\hspace*{10mm} $C$ \textit{Clipping Radius} \\
\hspace*{10mm} $\sigma$ \textit{Noise Multiplier} \\  
\hspace*{10mm} $\beta$ \textit{Momentum decay factor} \\
\hspace*{10mm} $\gamma$ \textit{Residual decay factor} \\
\hspace*{10mm} $\textrm{Compress}: \mathbb{R}^m \Rightarrow (\mathbb{R}^m, \mathbb{R}^m)$ \textit{Compression Function which outputs compressed input and residual} \\
\textbf{State:} $\textbf{v}_{sender}$ \textit{Velocity on sender device, initialized with zeros} \\
\hspace*{10mm} $\textbf{v}_{receiver}$ \textit{Velocity on receiver device, initialized with zeros} \\
\hspace*{10mm} $\textbf{r}$ \textit{Residual on server device, initialized with zeros} 
\begin{algorithmic}
\State $\textrm{Grad} \:\gets\: \{\:\nabla_{\theta}L(\textbf{X}_i, \textbf{y}_i)\:\}_{i=1}^B$
\State $\textrm{DPGrad} \:\gets\: \{\: \textrm{SphereProjection}_C(\textbf{g}_i) + \mathcal{N}(\textbf{0},\,\sigma C) \:|\: \textbf{g}_i \in \textrm{Grad} \:\}_{i=1}^B$ \Comment{Apply DP mechanism}
\State $\textrm{DCGrad} \:\gets\: \{\: \textrm{SphereProjection}_C(\textbf{g}_i) \:|\: \textbf{g}_i \in \textrm{DPGrad} \:\}_{i=1}^B$ \Comment{Apply second clipping}

\State $\textbf{v}_{sender} \:\gets\: \beta \textbf{v}_{sender} + (1-\beta) \frac{1}{|\textrm{DCGrad}|} \sum_{\textbf{g} \in \textrm{DCGrad}} \textbf{g}$ \Comment{Compute new velocity}
\State $\textbf{a} \:\gets\: \textbf{v}_{sender} - \textbf{v}$ \Comment{Compute acceleration}

\State $(\textbf{v}_{compressed}, \textbf{r}_v) \:\gets\: \textrm{Compress}(\textbf{v} + \gamma \textbf{r})$ \Comment{Compress velocity and compute the residual}
\State $(\textbf{a}_{compressed}, \textbf{r}_a) \:\gets\: \textrm{Compress}(\textbf{a} + \gamma \textbf{r})$ \Comment{Compress acceleration and compute the residual}

\If{$||\textbf{r}_v||_2 \le ||\textbf{r}_a||_2$} \Comment{Compare error between velocity and acceleration}
    \State $\textbf{r} \:\gets\: \textbf{r}_v$ \Comment{Update the residual}
    \State $\textbf{v}_{receiver} \:\gets\: \textbf{v}_{compressed}$ \Comment{Update the client using compressed velocity}
\Else
    \State $\textbf{r} \:\gets\: \textbf{r}_a$ \Comment{Update the residual}
    \State $\textbf{v}_{receiver} \:\gets\: \textbf{v}_{receiver} + \textbf{a}_{compressed}$ \Comment{Update the client using compressed acceleration}
\EndIf
\end{algorithmic}
\end{algorithm}

\section{Conclusion}
Our work demonstrates the interaction between differential privacy mechanisms and gradient compression and their combined effect on accuracy in deep learning models. We observe that gradient compression has a tendency to decrease the model's sensitivity to noise and sometimes undo the negative impacts of noise on accuracy. We explain this through analyzing the gradient error distribution between bias and variance. We apply this reasoning by proposing methods of better optimizing the bias-variance trade-off in differentially private learning.

We follow this study with a recommendation on how to improve test accuracy under the context of differentially private deep learning and gradient compression. We evaluate this proposal and find that it can reduce the negative impact of noise added by differential privacy mechanisms  on test accuracy by up to 24.6\%, and reduce the negative impact of gradient sparsification on test accuracy by up to 15.1\%.
  \appendix
    \chapter{Test Accuracy by Privacy Bound and Bandwidth}

\begin{figure}[h]
    \centering
    \includegraphics[width=1.0\linewidth]{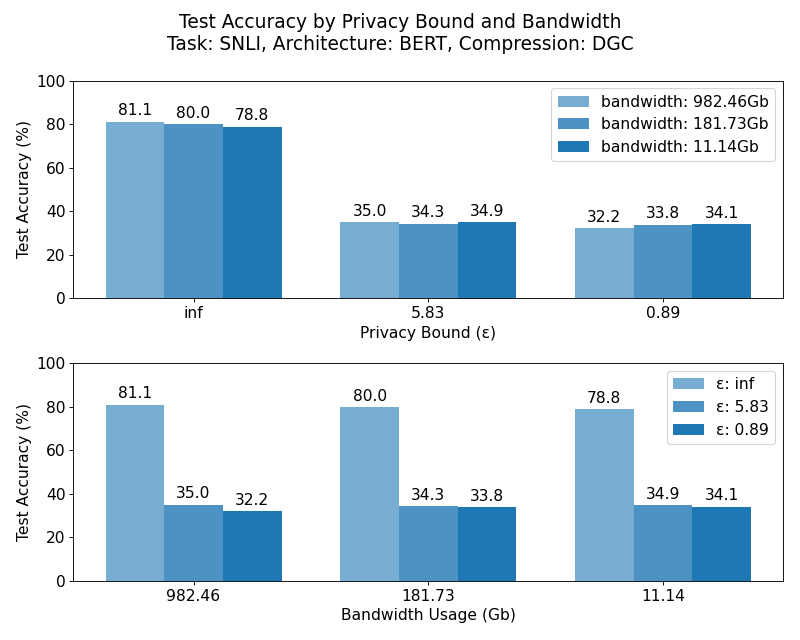}
    \caption{Test accuracy after 1 epoch of training the SNLI task using DGC. (Top) grouped by differential privacy bound ($\epsilon$). (Bottom) grouped by upstream network usage (Gb).}
    \label{fig:skyline_bert_local_dgc}
\end{figure}
\begin{figure}[h]
    \centering
    \includegraphics[width=1.0\linewidth]{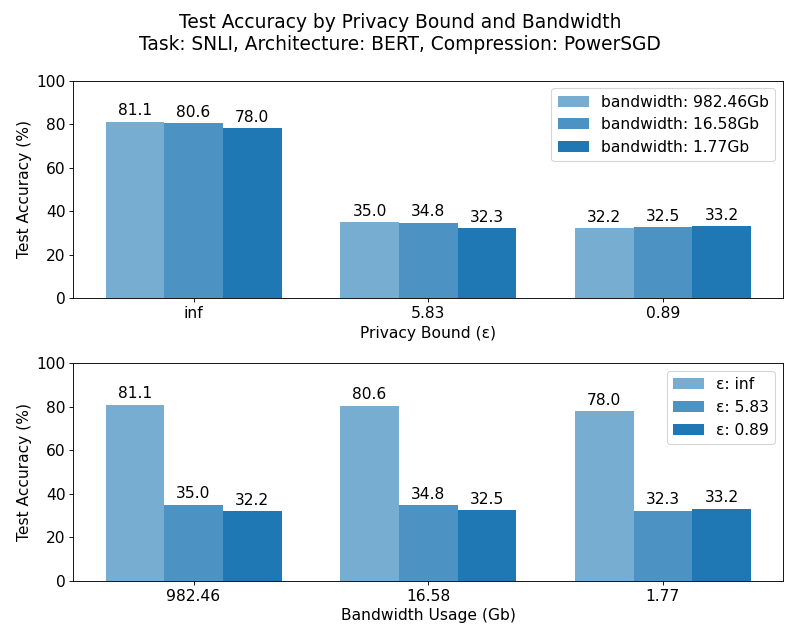}
    \caption{Test accuracy after 1 epoch of training the SNLI task using PowerSGD. (Top) grouped by differential privacy bound ($\epsilon$). (Bottom) grouped by upstream network usage (Gb).}
    \label{fig:skyline_bert_local_power_sgd}
\end{figure}
\begin{figure}[h]
    \centering
    \includegraphics[width=1.0\linewidth]{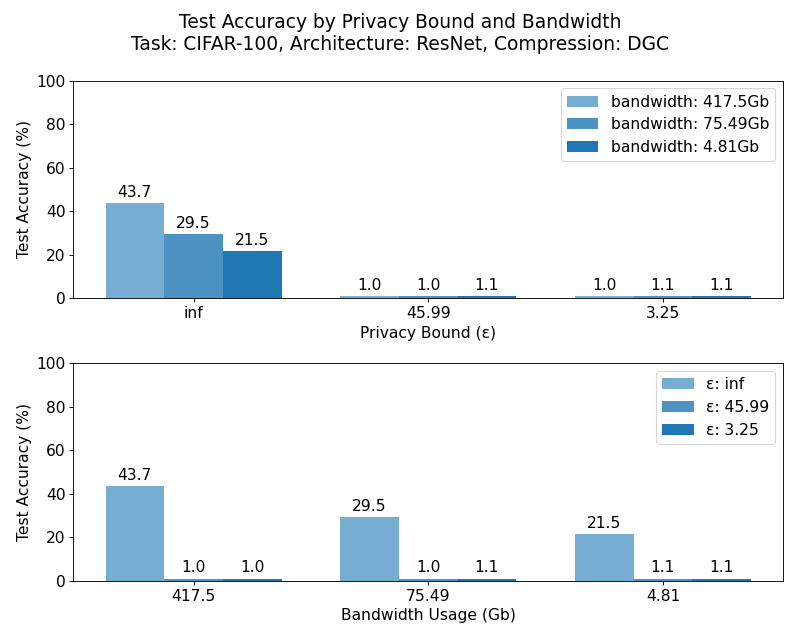}
    \caption{Test accuracy averaged over last 10 epochs after 1 epoch of training the CIFAR-100 task using DGC. (Top) grouped by differential privacy bound ($\epsilon$). (Bottom) grouped by upstream network usage (Gb).}
    \label{fig:skyline_resnet_local_dgc}
\end{figure}
\begin{figure}[h]
    \centering
    \includegraphics[width=1.0\linewidth]{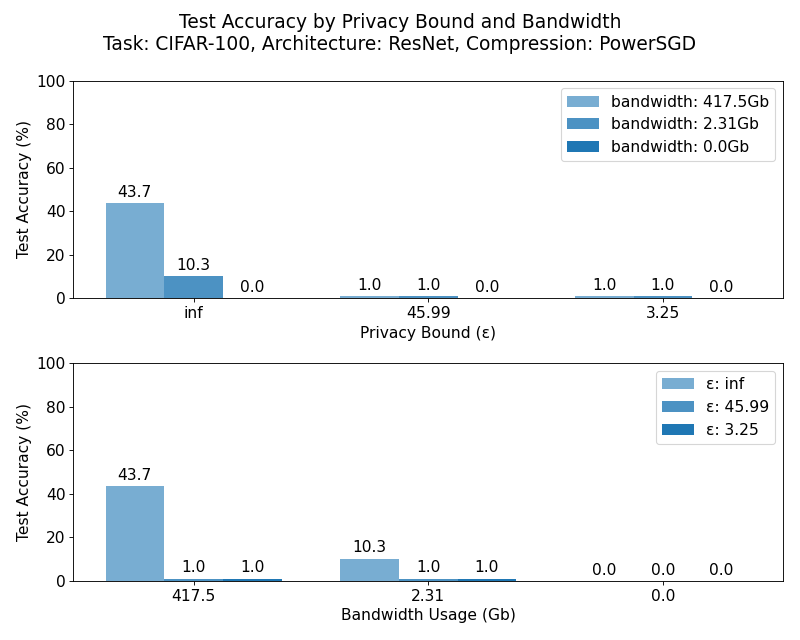}
    \caption{Test accuracy averaged over last 10 epochs after 1 epoch of training the CIFAR-100 task using PowerSGD. (Top) grouped by differential privacy bound ($\epsilon$). (Bottom) grouped by upstream network usage (Gb).}
    \label{fig:skyline_resnet_local_power_sgd}
\end{figure}

\chapter{Test Accuracy vs Epochs}
\begin{figure}[h]
    \centering
    \includegraphics[width=1.0\linewidth]{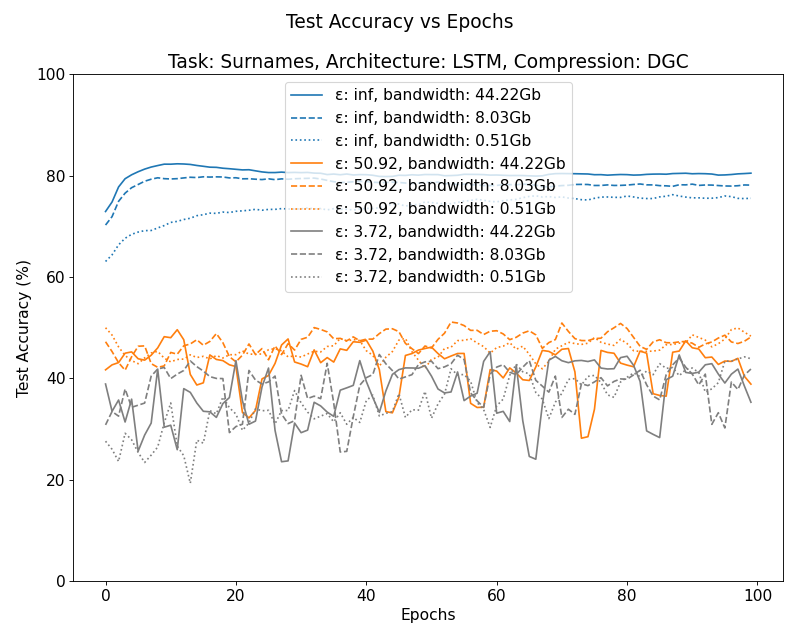}
    \caption{Test Accuracy over 100 Epochs of the Surnames task.}
    \label{fig:timeline_lstm_local_dgc}
\end{figure}
\begin{figure}[h]
    \centering
    \includegraphics[width=1.0\linewidth]{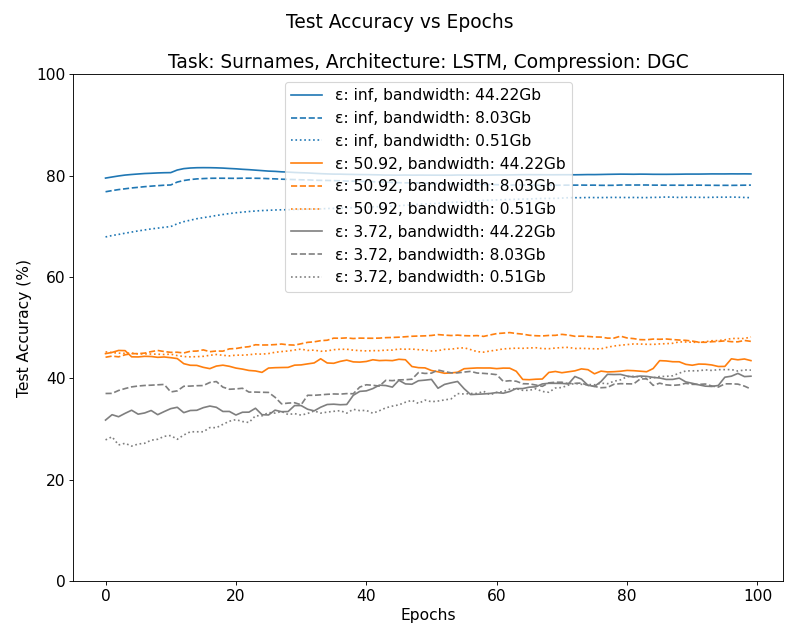}
    \caption{Smoothed test Accuracy over 100 Epochs of the Surnames task. Smoothing is done using a 1D averaging convolution with window size 20 and no padding.}
    \label{fig:timeline_lstm_local_dgc_smooth}
\end{figure}
\begin{figure}[h]
    \centering
    \includegraphics[width=1.0\linewidth]{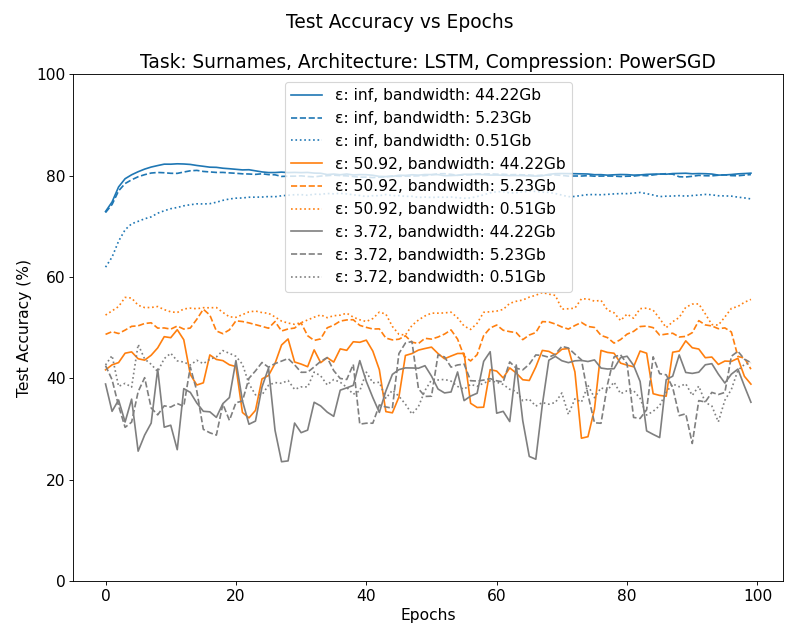}
    \caption{Test Accuracy over 100 Epochs of the Surnames task.}
    \label{fig:timeline_lstm_local_power_sgd}
\end{figure}
\begin{figure}[h]
    \centering
    \includegraphics[width=1.0\linewidth]{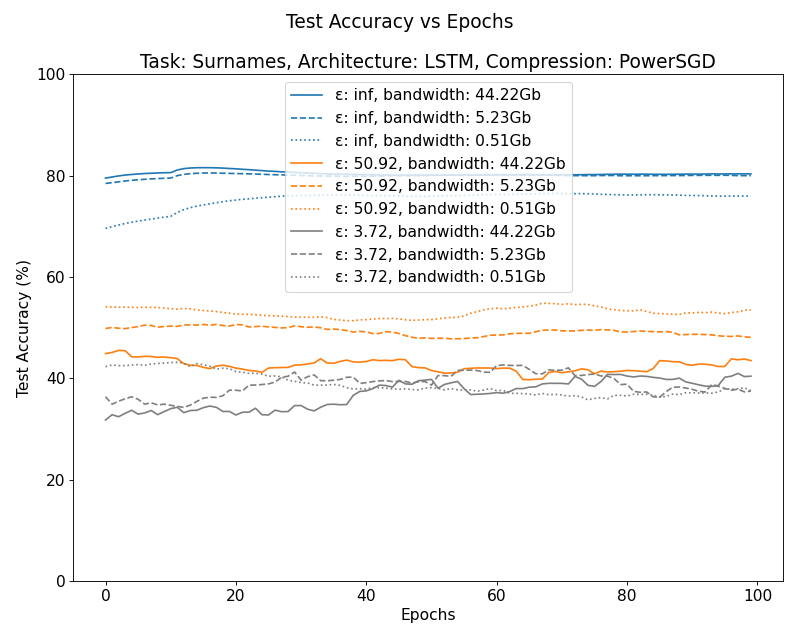}
    \caption{Smoothed test Accuracy over 100 Epochs of the Surnames task. Smoothing is done using a 1D averaging convolution with window size 20 and no padding.}
    \label{fig:timeline_lstm_local_power_sgd_smooth}
\end{figure}

\begin{figure}[h]
    \centering
    \includegraphics[width=1.0\linewidth]{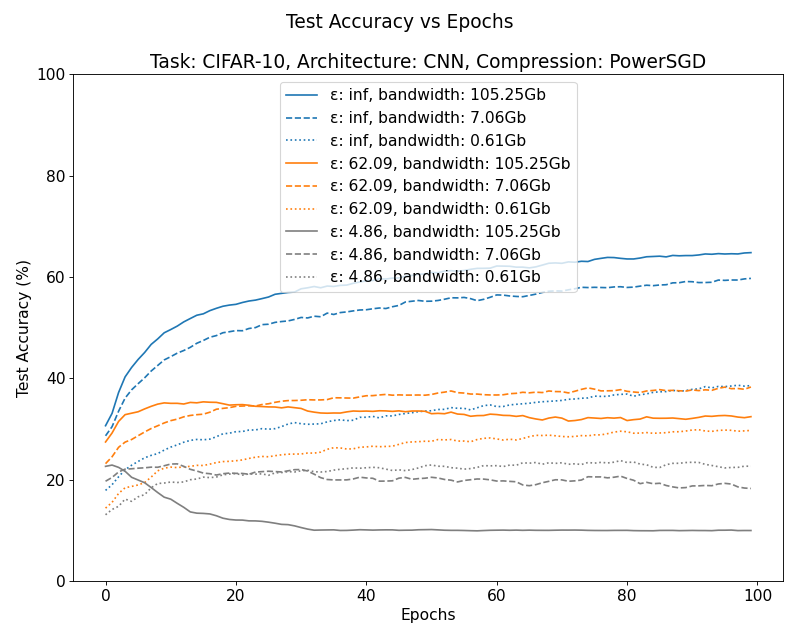}
    \caption{Test Accuracy over 100 Epochs of the CIFAR-10 task.}
    \label{fig:timeline_cnn_local_power_sgd}
\end{figure}
\begin{figure}[h]
    \centering
    \includegraphics[width=1.0\linewidth]{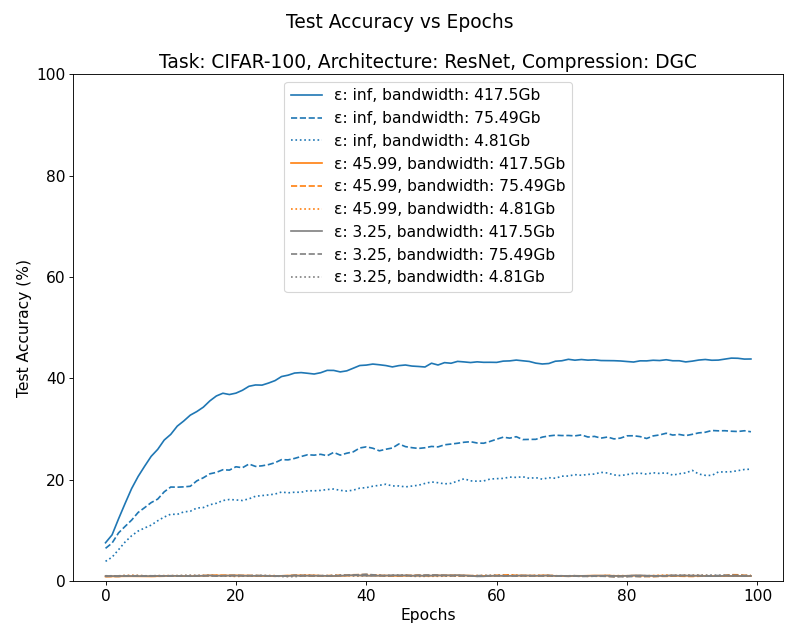}
    \caption{Test Accuracy over 100 Epochs of the CIFAR-100 task.}
    \label{fig:timeline_resnet_local_dgc}
\end{figure}
\begin{figure}[h]
    \centering
    \includegraphics[width=1.0\linewidth]{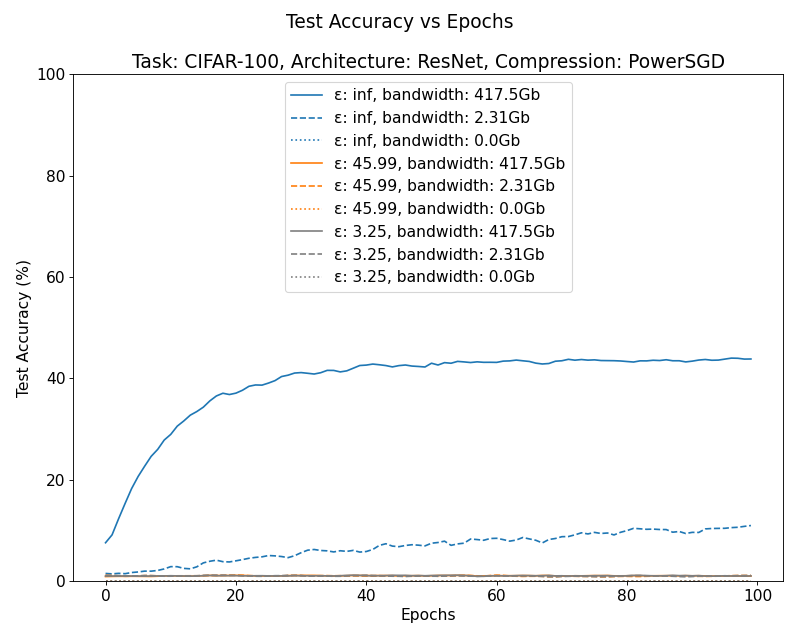}
    \caption{Test Accuracy over 100 Epochs of the CIFAR-100 task.}
    \label{fig:timeline_resnet_local_power_sgd}
\end{figure}

\chapter{Test Accuracy vs Gradient Error}
\begin{figure}[h]
    \centering
    \includegraphics[width=1.0\linewidth]{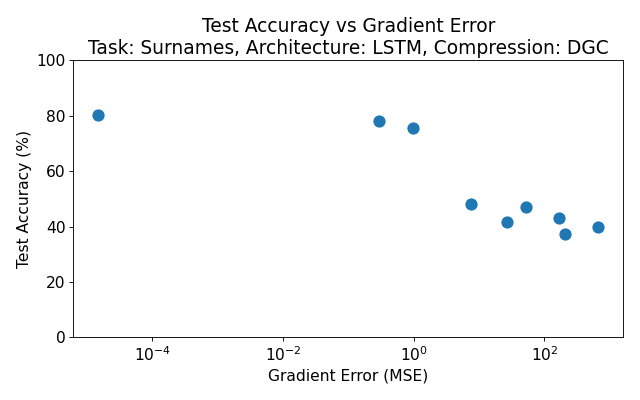}
    \caption{Test accuracy vs gradient error (log base-10 scale) for the Surnames task with DGC.}
    \label{fig:explanation_lstm_local_dgc}
\end{figure}
\begin{figure}[h]
    \centering
    \includegraphics[width=1.0\linewidth]{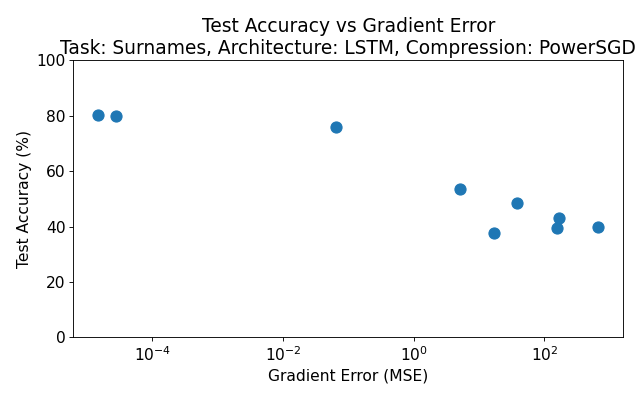}
    \caption{Test accuracy vs gradient error (log base-10 scale) for the Surnames task with PowerSGD.}
    \label{fig:explanation_lstm_local_power_sgd}
\end{figure}
\begin{figure}[h]
    \centering
    \includegraphics[width=1.0\linewidth]{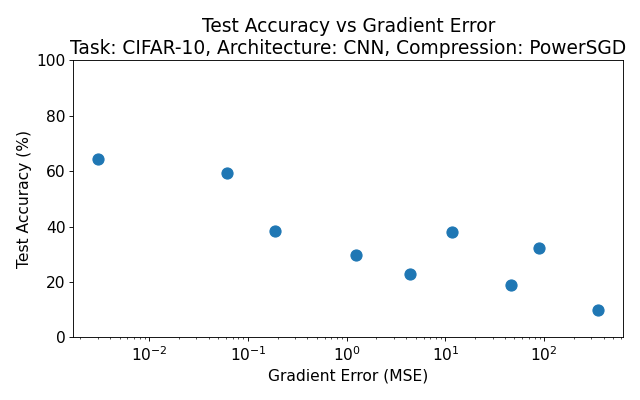}
    \caption{Test accuracy vs gradient error (log base-10 scale) for the CIFAR-10 task with PowerSGD.}
    \label{fig:explanation_cnn_local_power_sgd}
\end{figure}
\begin{figure}[h]
    \centering
    \includegraphics[width=1.0\linewidth]{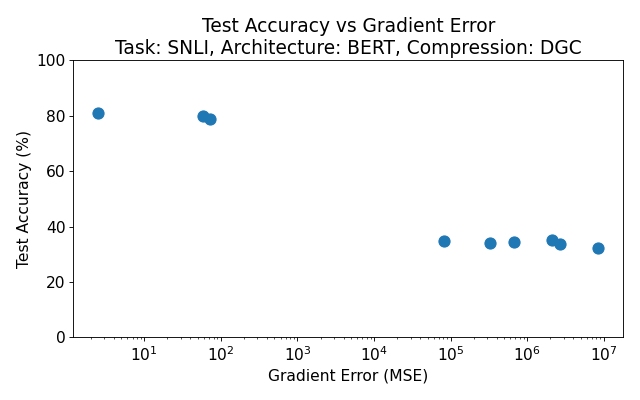}
    \caption{Test accuracy vs gradient error (log base-10 scale) for the SNLI task with DGC.}
    \label{fig:explanation_bert_local_dgc}
\end{figure}
\begin{figure}[h]
    \centering
    \includegraphics[width=1.0\linewidth]{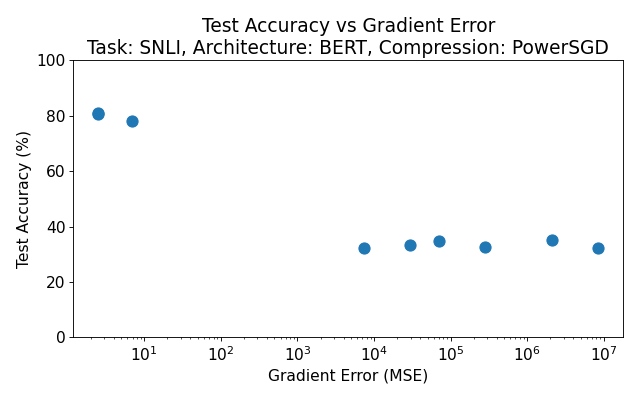}
    \caption{Test accuracy vs gradient error (log base-10 scale) for the SNLI task with PowerSGD.}
    \label{fig:explanation_bert_local_power_sgd}
\end{figure}
\begin{figure}[h]
    \centering
    \includegraphics[width=1.0\linewidth]{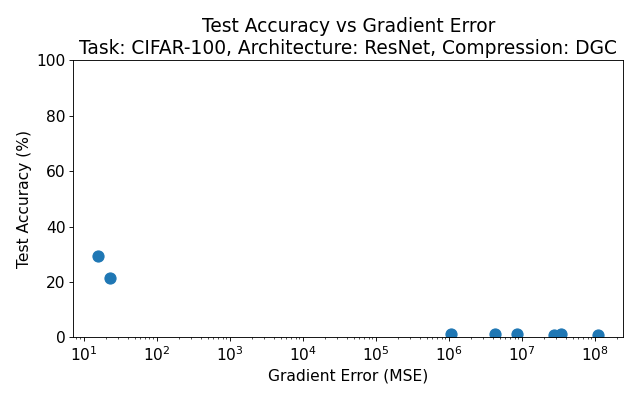}
    \caption{Test accuracy vs gradient error (log base-10 scale) for the CIFAR-100 task with DGC.}
    \label{fig:explanation_resnet_local_dgc}
\end{figure}
\begin{figure}[h]
    \centering
    \includegraphics[width=1.0\linewidth]{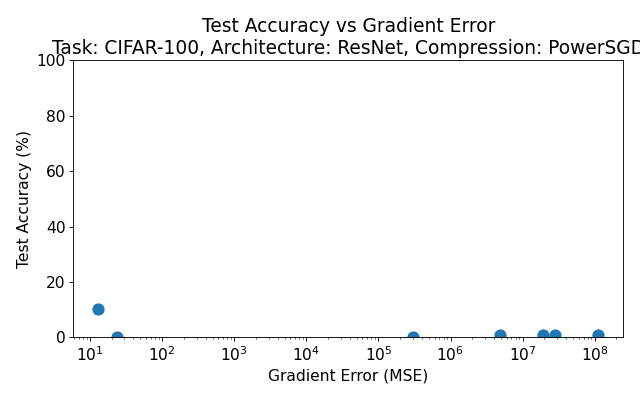}
    \caption{Test accuracy vs gradient error (log base-10 scale) for the CIFAR-100 task with PowerSGD.}
    \label{fig:explanation_resnet_local_power_sgd}
\end{figure}

\chapter{Gradient Error vs Privacy Bound and Bandwidth}
\begin{figure}[h]
    \centering
    \includegraphics[width=1.0\linewidth]{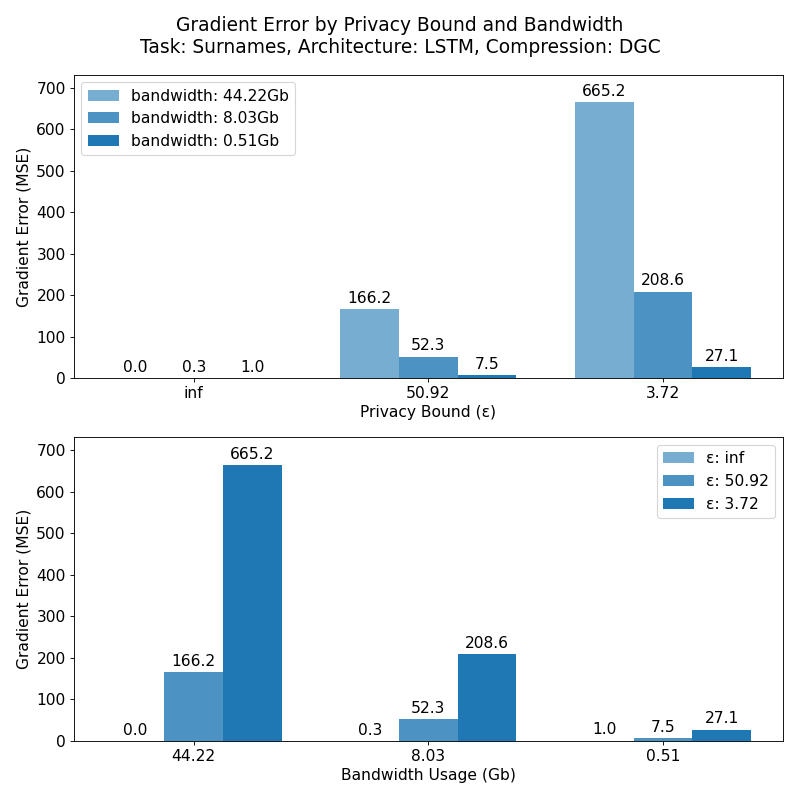}
    \caption{Gradient Error vs Privacy Bound ($\epsilon$) and Bandwidth (Gb) for the Surnames task with DGC.}
    \label{fig:error_skyline_lstm_local_dgc}
\end{figure}
\begin{figure}[h]
    \centering
    \includegraphics[width=1.0\linewidth]{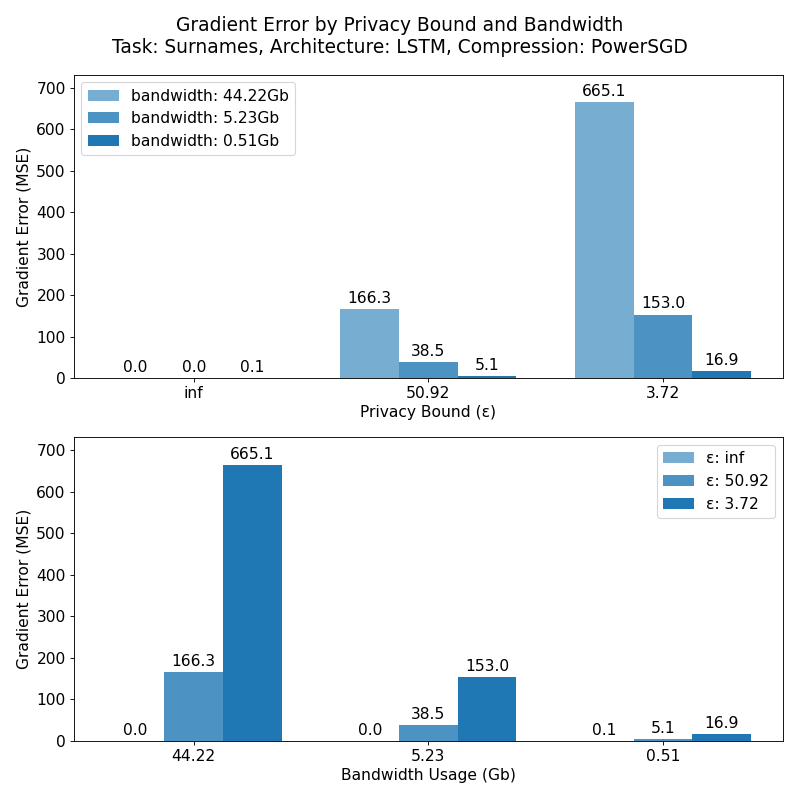}
    \caption{Gradient Error vs Privacy Bound ($\epsilon$) and Bandwidth (Gb) for the Surnames task with PowerSGD.}
    \label{fig:error_skyline_lstm_local_power_sgd}
\end{figure}
\begin{figure}[h]
    \centering
    \includegraphics[width=1.0\linewidth]{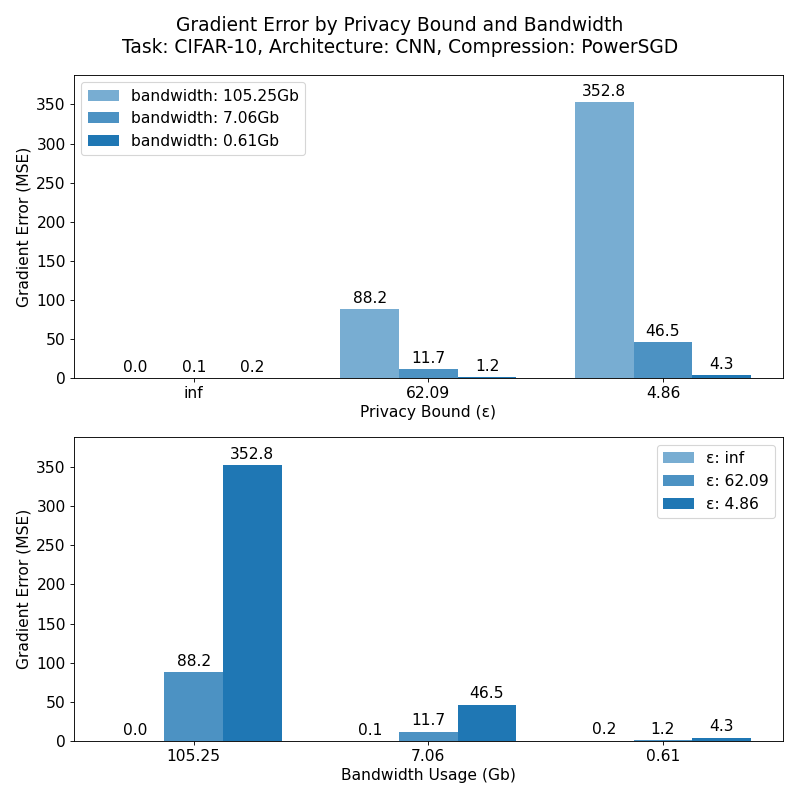}
    \caption{Gradient Error vs Privacy Bound ($\epsilon$) and Bandwidth (Gb) for the CIFAR-10 task with PowerSGD.}
    \label{fig:error_skyline_cnn_local_power_sgd}
\end{figure}
\begin{figure}[h]
    \centering
    \includegraphics[width=1.0\linewidth]{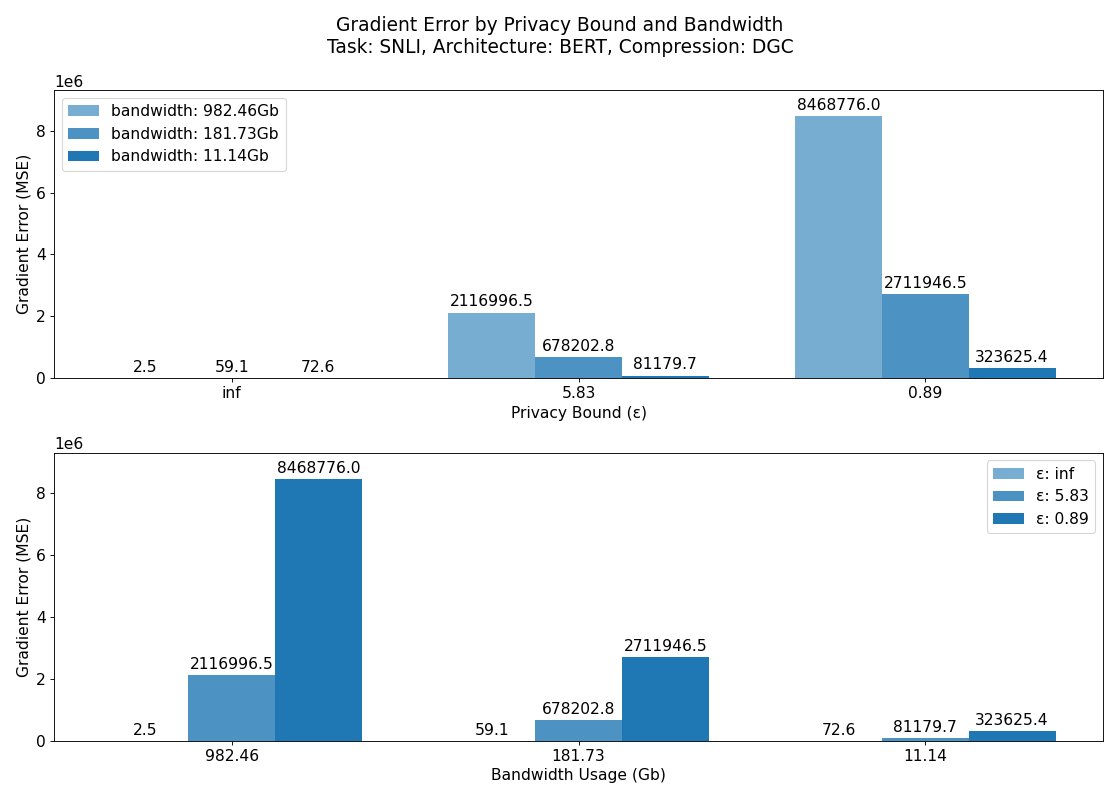}
    \caption{Gradient Error vs Privacy Bound ($\epsilon$) and Bandwidth (Gb) for the SNLI task with DGC.}
    \label{fig:error_skyline_bert_local_dgc}
\end{figure}
\begin{figure}[h]
    \centering
    \includegraphics[width=1.0\linewidth]{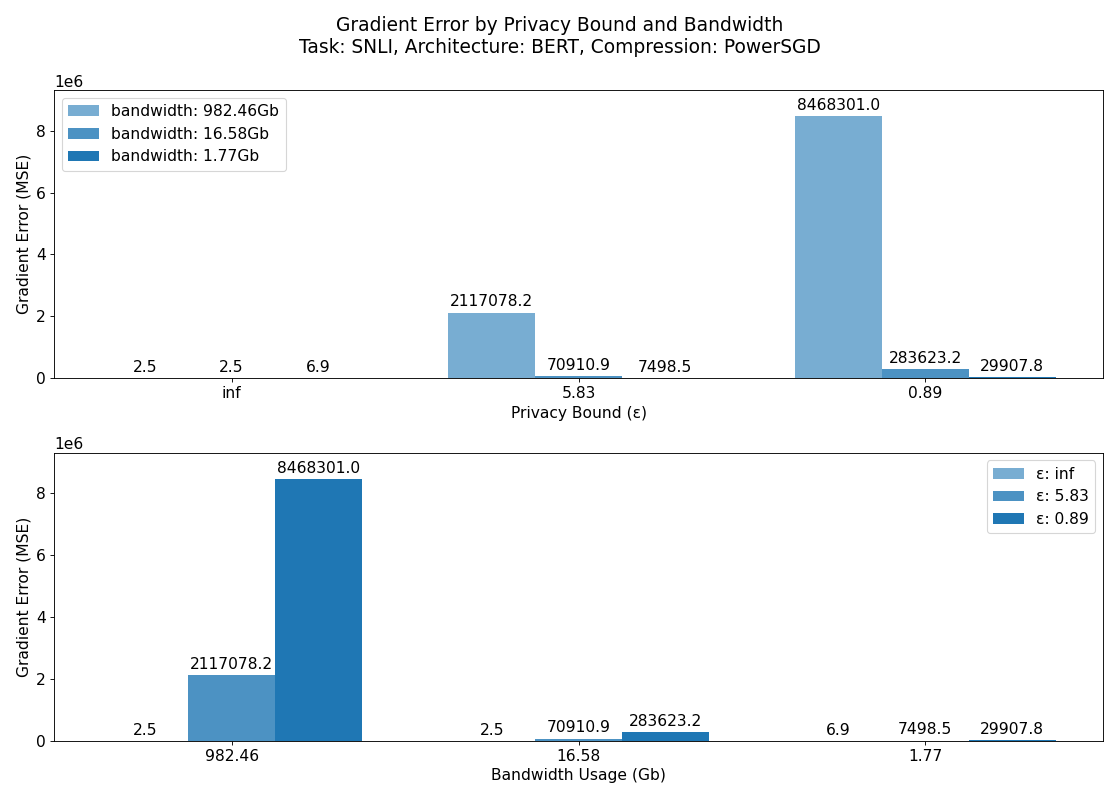}
    \caption{Gradient Error vs Privacy Bound ($\epsilon$) and Bandwidth (Gb) for the SNLI task with PowerSGD.}
    \label{fig:error_skyline_bert_local_power_sgd}
\end{figure}
\begin{figure}[h]
    \centering
    \includegraphics[width=1.0\linewidth]{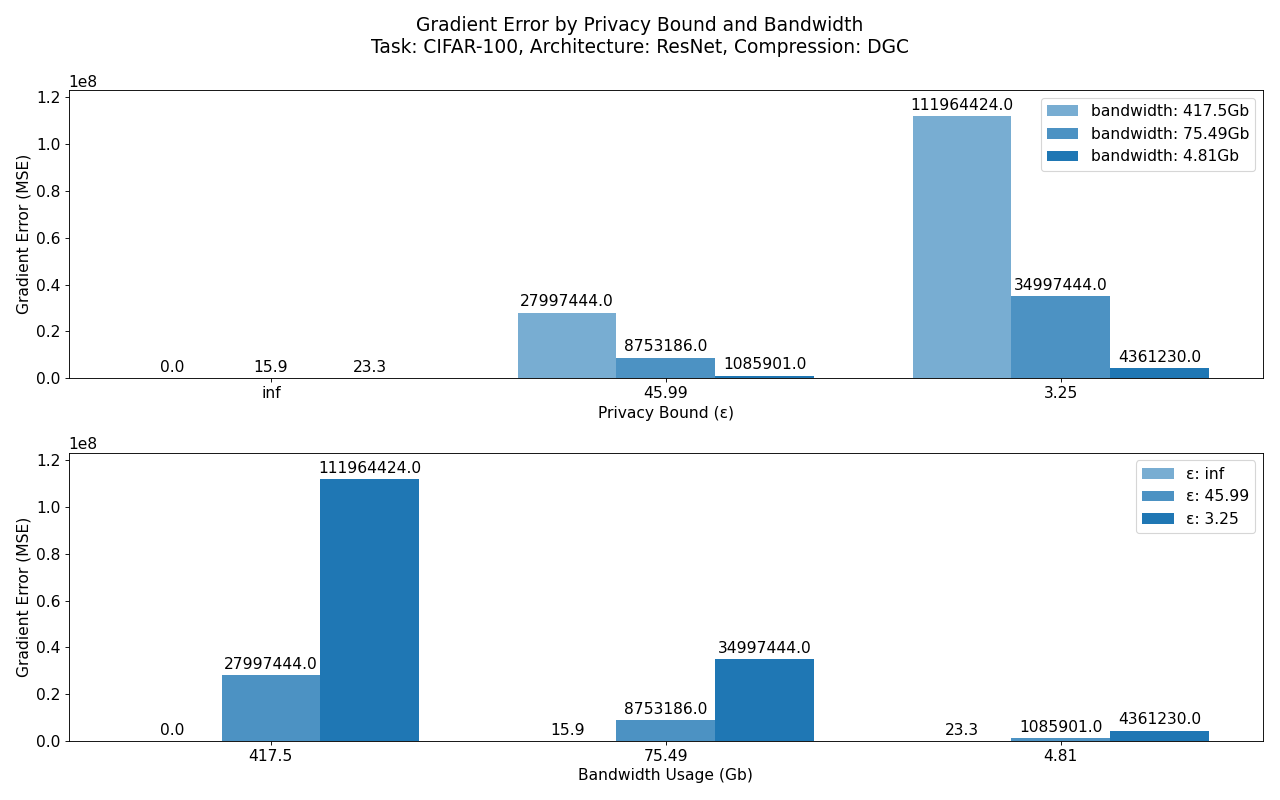}
    \caption{Gradient Error vs Privacy Bound ($\epsilon$) and Bandwidth (Gb) for the CIFAR-100 task with DGC.}
    \label{fig:error_skyline_resnet_local_dgc}
\end{figure}
\begin{figure}[h]
    \centering
    \includegraphics[width=1.0\linewidth]{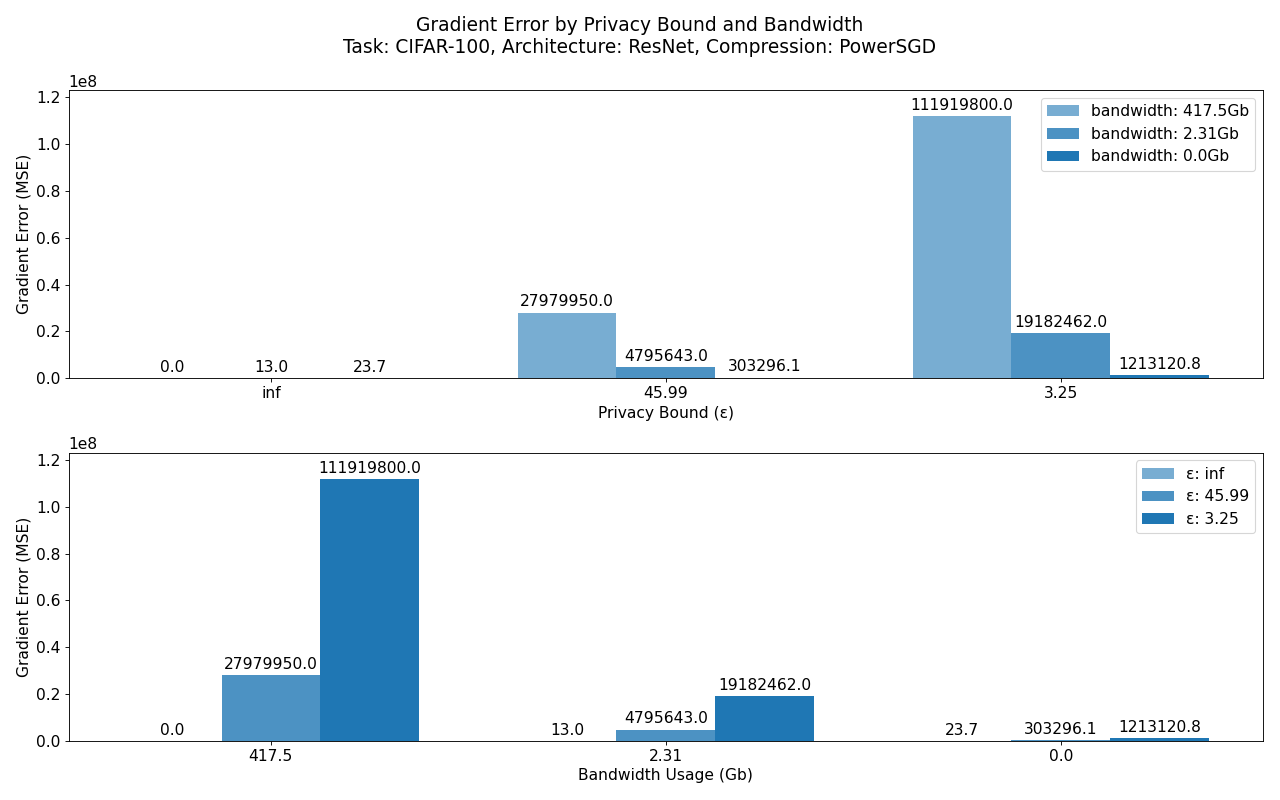}
    \caption{Gradient Error vs Privacy Bound ($\epsilon$) and Bandwidth (Gb) for the CIFAR-100 task with PowerSGD.}
    \label{fig:error_skyline_resnet_local_power_sgd}
\end{figure}

\chapter{Gradient Error Breakdown}
\begin{figure}[h]
    \centering
    \includegraphics[width=1.0\linewidth]{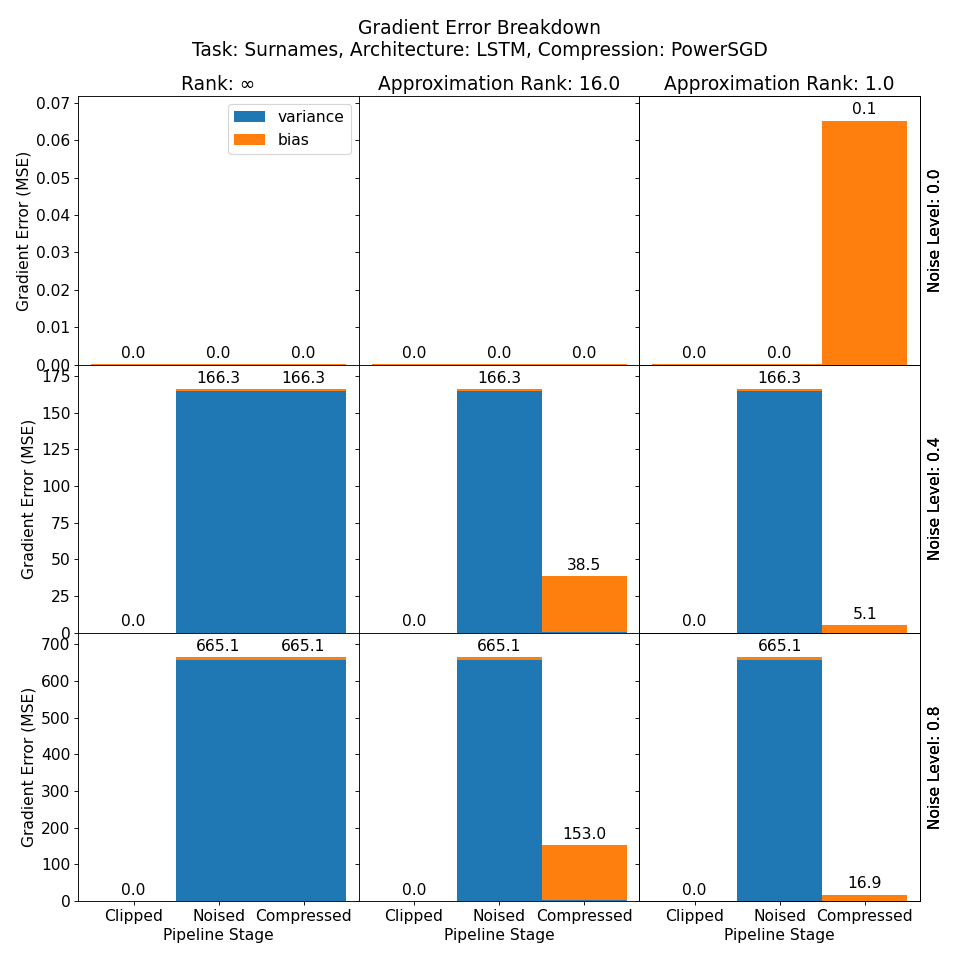}
    \caption{Breakdown of gradient error into bias and variance at different stages (after clipping, after noising, and after compression) for the Surnames task with PowerSGD.}
    \label{fig:error_breakdown_lstm_local_power_sgd}
\end{figure}
\begin{figure}[h]
    \centering
    \includegraphics[width=1.0\linewidth]{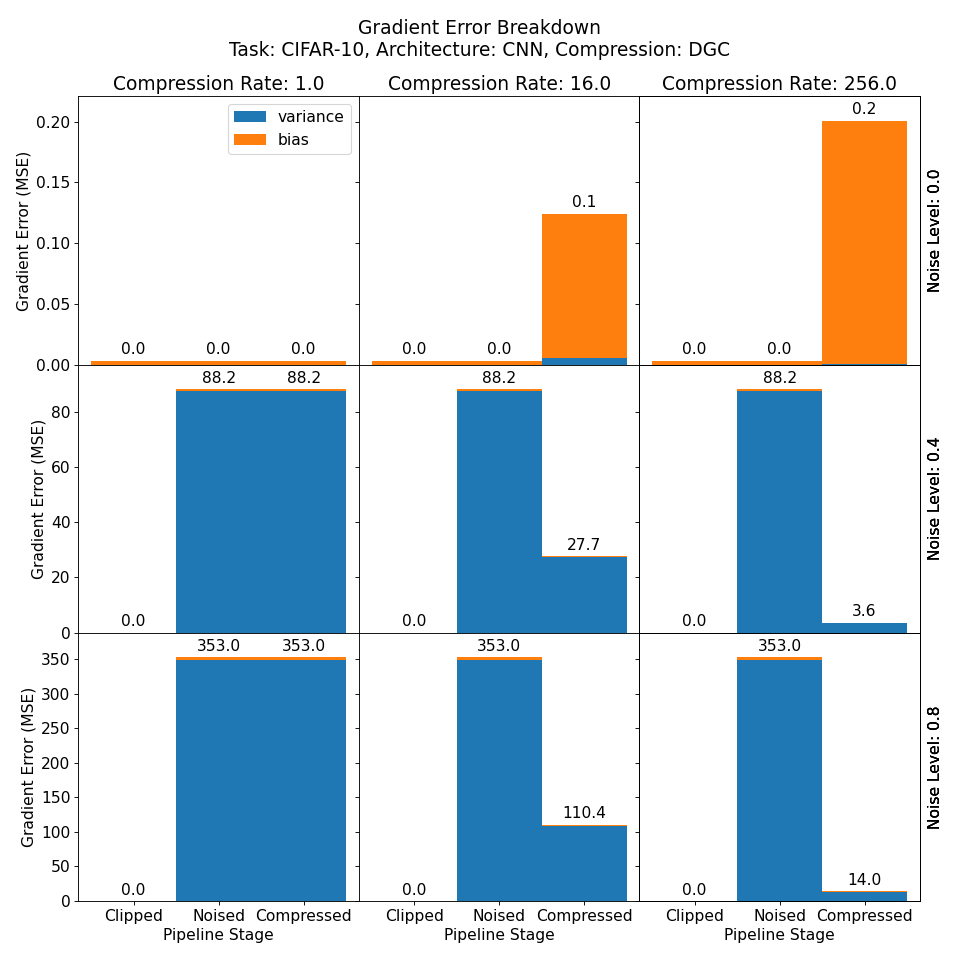}
    \caption{Breakdown of gradient error into bias and variance at different stages (after clipping, after noising, and after compression) for the CIFAR-10 task with DGC.}
    \label{fig:error_breakdown_cnn_local_dgc}
\end{figure}
\begin{figure}[h]
    \centering
    \includegraphics[width=1.0\linewidth]{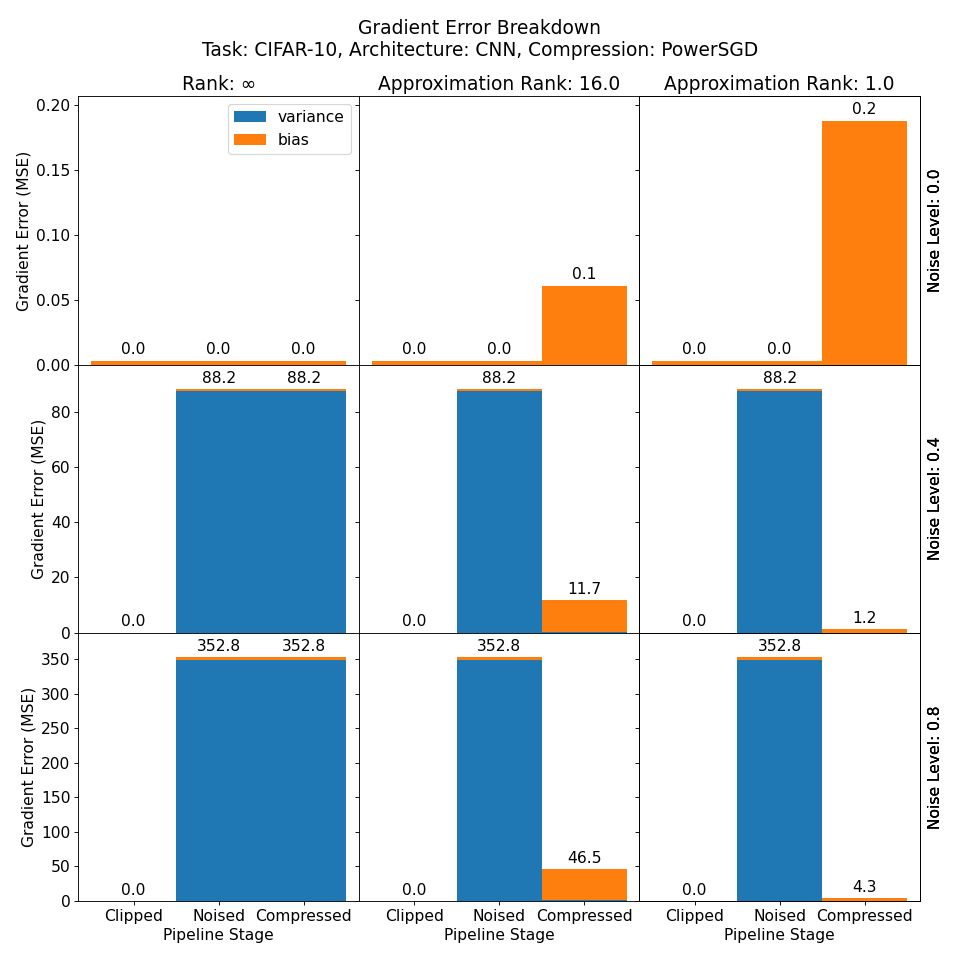}
    \caption{Breakdown of gradient error into bias and variance at different stages (after clipping, after noising, and after compression) for the CIFAR-10 task with PowerSGD.}
    \label{fig:error_breakdown_cnn_local_power_sgd}
\end{figure}
\begin{figure}[h]
    \centering
    \includegraphics[width=1.0\linewidth]{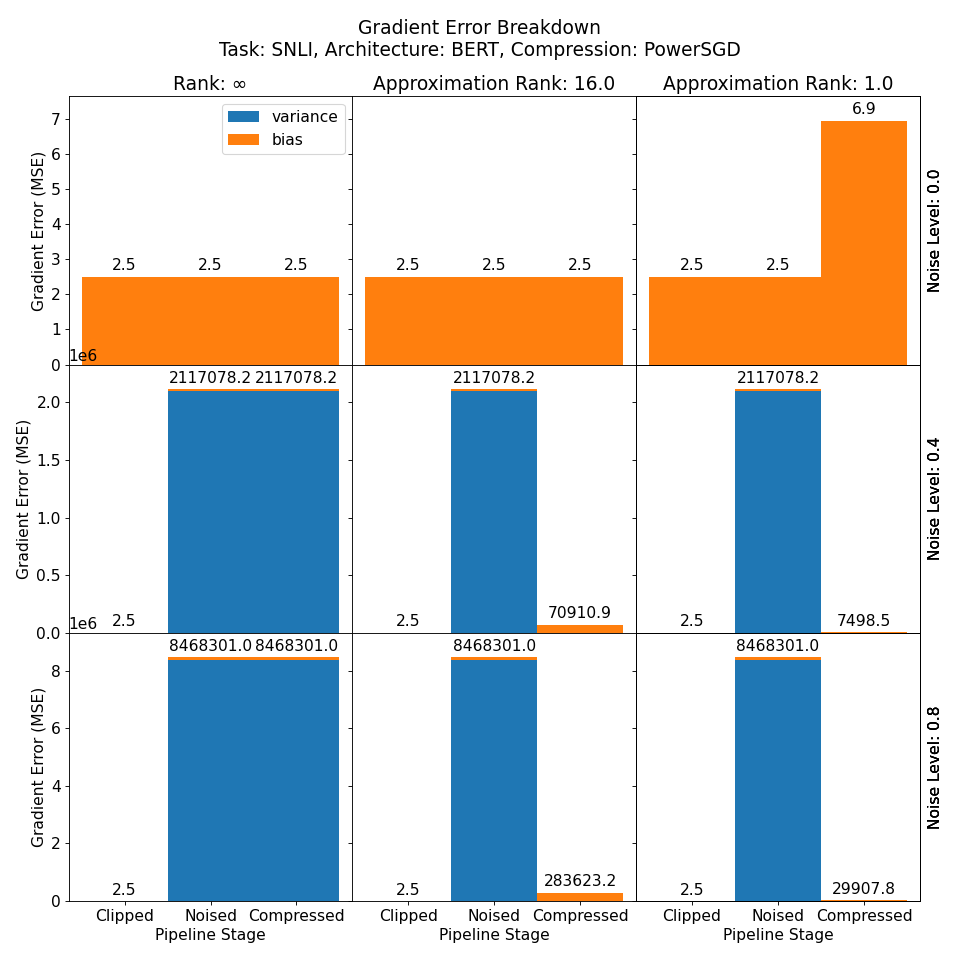}
    \caption{Breakdown of gradient error into bias and variance at different stages (after clipping, after noising, and after compression) for the SNLI task with PowerSGD.}
    \label{fig:error_breakdown_bert_local_power_sgd}
\end{figure}
\begin{figure}[h]
    \centering
    \includegraphics[width=1.0\linewidth]{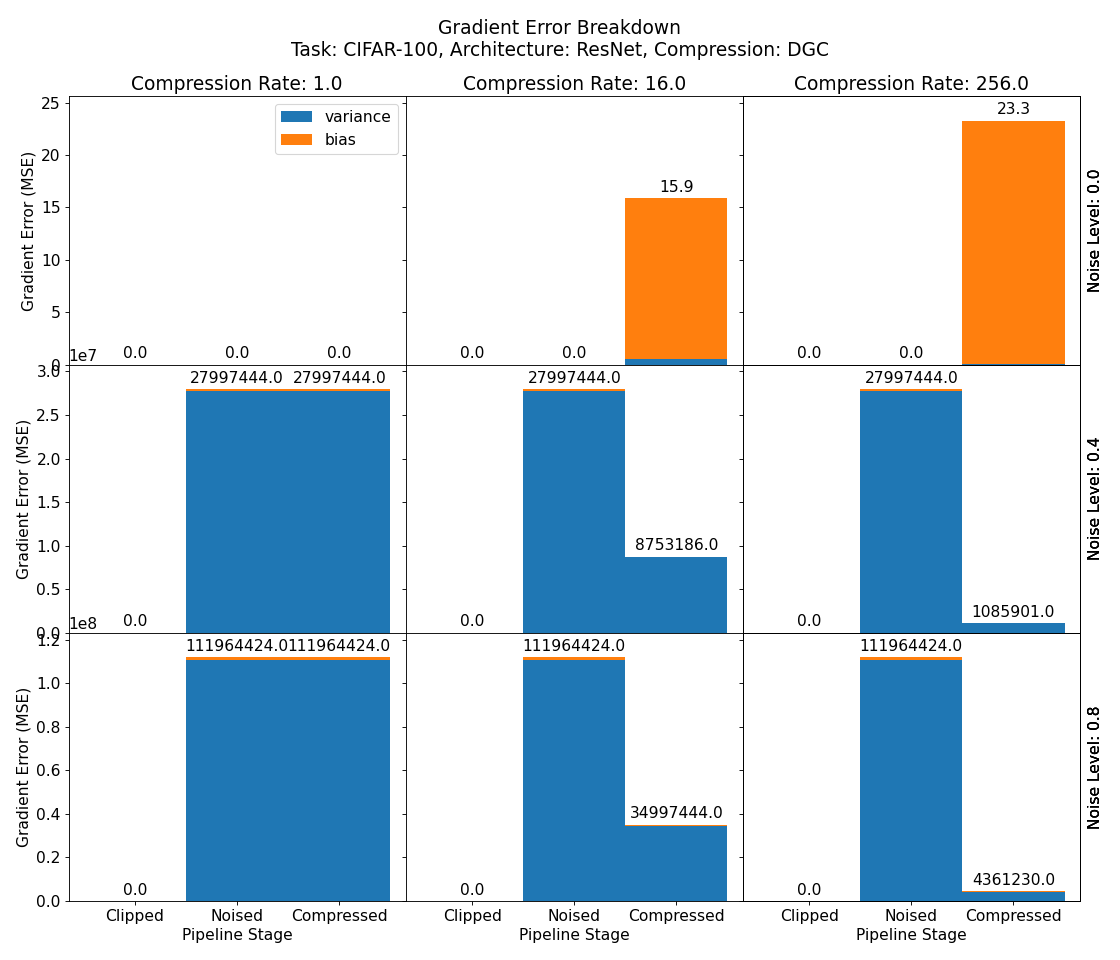}
    \caption{Breakdown of gradient error into bias and variance at different stages (after clipping, after noising, and after compression) for the CIFAR-100 task with DGC.}
    \label{fig:error_breakdown_resnet_local_dgc}
\end{figure}
\begin{figure}[h]
    \centering
    \includegraphics[width=1.0\linewidth]{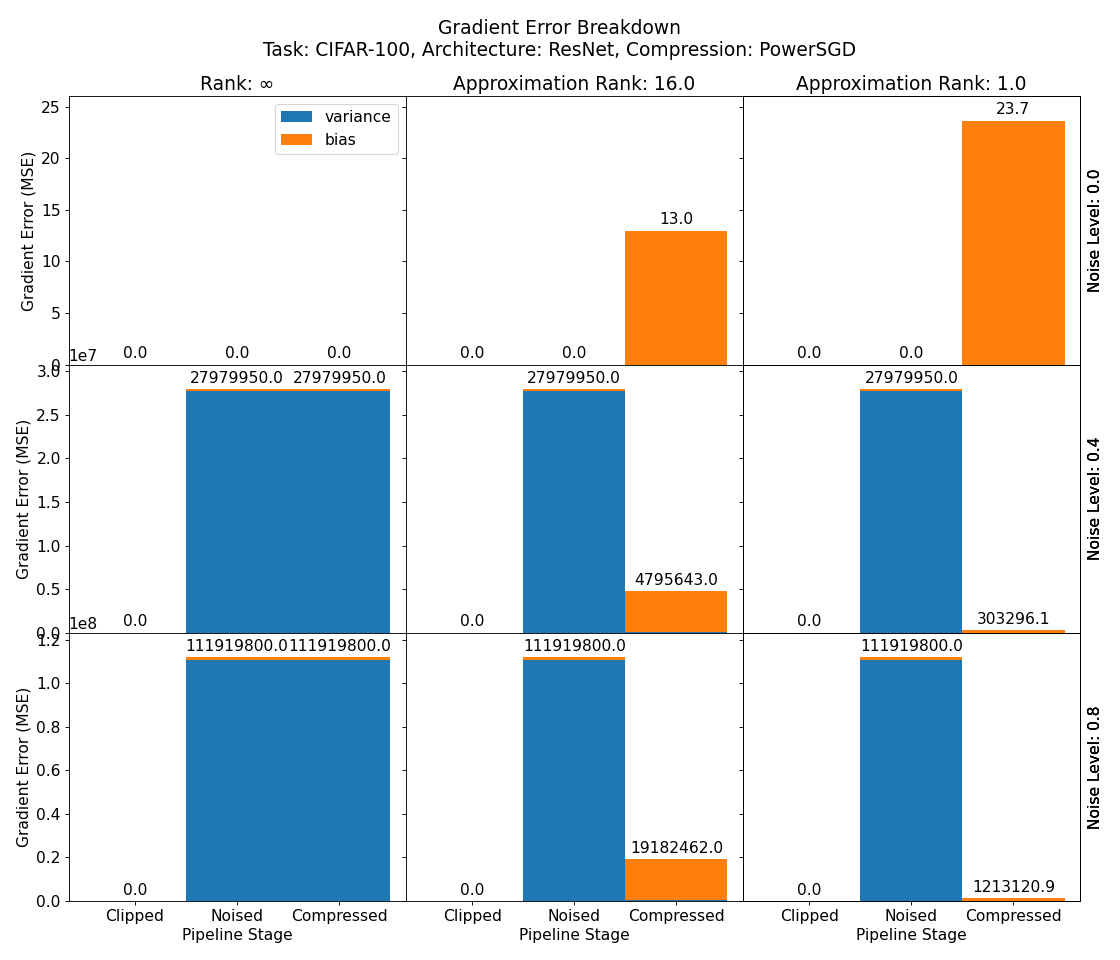}
    \caption{Breakdown of gradient error into bias and variance at different stages (after clipping, after noising, and after compression) for the CIFAR-100 task with PowerSGD.}
    \label{fig:error_breakdown_resnet_local_power_sgd}
\end{figure}

\chapter{Minimal-Error Clipping}
\begin{figure}[h]
    \centering
    \includegraphics[width=1.0\linewidth]{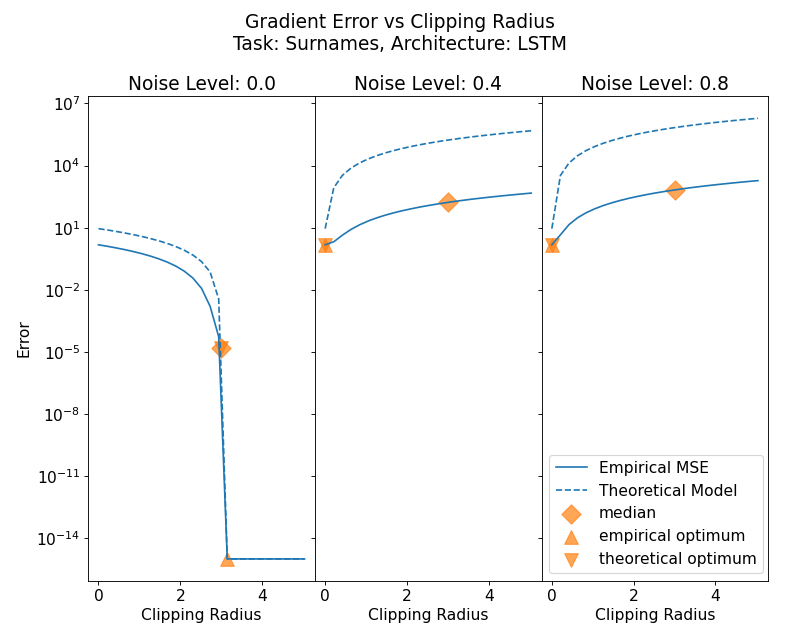}
    \caption{Relationship between gradient error and clipping value for Surnames task.}
    \label{fig:optimal_clipping_lstm}
\end{figure}
\begin{figure}[h]
    \centering
    \includegraphics[width=1.0\linewidth]{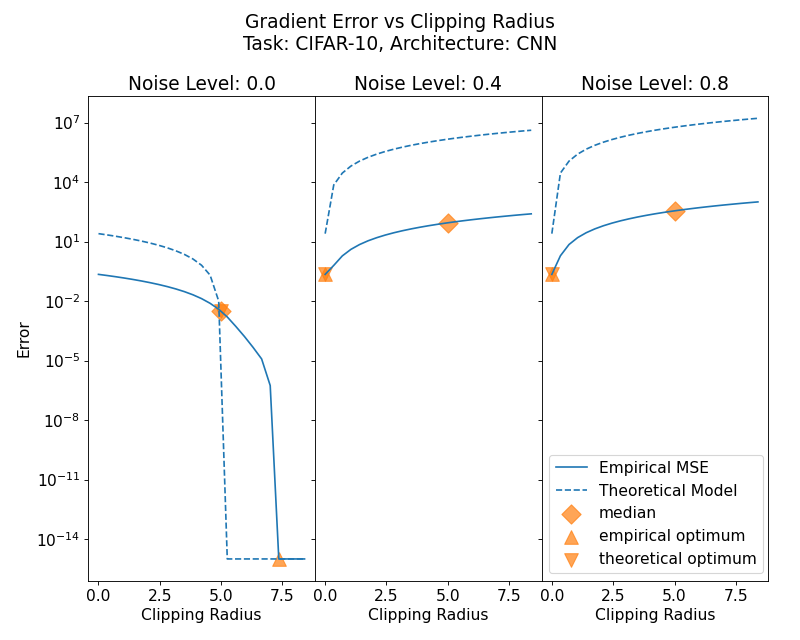}
    \caption{Relationship between gradient error and clipping value for CIFAR-10 task.}
    \label{fig:optimal_clipping_cnn}
\end{figure}
  \backmatter
  \printbibliography[heading=bibintoc]
\end{document}